\documentclass[11pt]{article}

\usepackage[final]{acl}

\usepackage{times}
\usepackage{latexsym}
\usepackage{multirow}
\usepackage{xcolor}
\usepackage{amsmath}
\usepackage{amsfonts}
\usepackage{amssymb}
\usepackage{colortbl}
\usepackage{booktabs}
\usepackage{subfigure}
\usepackage{algorithm}
\usepackage{algorithmic}
\usepackage{newfloat}
\usepackage{listings}
\usepackage{url}

\usepackage[T1]{fontenc}

\usepackage[utf8]{inputenc}

\usepackage{microtype}

\usepackage{inconsolata}

\usepackage{graphicx}

\title{HiPrune: Hierarchical Attention for Efficient Token Pruning in Vision-Language Models}

\author{Jizhihui Liu\textsuperscript{1}$^{\footnotemark[1]}$, Guangdao Zhu\textsuperscript{1}$^{\footnotemark[1]}$, Feiyi Du\textsuperscript{1}$^{\footnotemark[1]}$, Niu Lian\textsuperscript{1}, Jun Li\textsuperscript{1,2} \\ {\bf Bin Chen\textsuperscript{1,2}$^{\footnotemark[2]}$}, {\bf Weili Guan\textsuperscript{1,2}}, {\bf Yaowei Wang\textsuperscript{1,2}} \\ \textsuperscript{1}Harbin Institute of Technology, Shenzhen \textsuperscript{2}Pengcheng Laboratory \\ danielement321@gmail.com, chenbin2021@hit.edu.cn}

\begin{document}
\maketitle
\footnotetext[1]{Equal Contribution.}
\footnotetext[2]{Corresponding Author.}

\begin{abstract}
Vision-Language Models (VLMs) encode images and videos into abundant tokens, which contain substantial redundancy and computation cost. While visual token pruning mitigates the issue, most existing methods lack insight into the intrinsic property of the vision encoder itself. In this work, we dive into the vision encoder and prove that the middle layers pay more attention to the main objects of the image qualitatively and quantitatively, while the deep layers to tokens with rich global information. Utilizing this \textbf{Hi}erarchical attention pattern, we propose \textbf{HiPrune}, a training-free and model-agnostic token \textbf{Prun}ing method. HiPrune identifies three types of visual tokens according to their attention in different phases of the vision encoder, which preserves different levels of information. By coupling with the similarity of text tokens, we propose a prompt-aware variance, \textbf{HiPrune++}, which further improves instruction following performance under a very low token budget. Extensive experiments across four representative VLMs show that HiPrune achieves up to 99.3\% of task accuracy with only 1/3 of the tokens, while reducing inference FLOPs by 58.7\%. HiPrune++ maintains up to 99.7\% accuracy with 2/9 tokens, highlighting robustness under high-resolution. Our code is available at \url{https://github.com/Danielement321/HiPrune}.
\end{abstract}

\section{Introduction}
Built on the success of Large Language Models (LLMs) \cite{touvron2023llama, yang2025qwen3}, Vision-Language Models (VLMs) \cite{team2023gemini, hurst2024gpt, liang2024survey} have demonstrated considerable capacity in various visual tasks. VLMs commonly comprise a vision encoder \cite{radford2021learning, zhai2023sigmoid}, an adaptor, and an LLM. The vision encoder is a vision transformer (ViT) \cite{dosovitskiy2020image, vaswani2017attention} that encodes the image into numerous tokens, which account for the biggest proportion of inputs and cause significant latency and memory demands. In LLaVA-1.5 \cite{liu2023visual, liu2024improved}, an image is encoded into 576 tokens, much longer than its textual counterparts \cite{zhang2025vispruner}. For VLMs that incorporate a native dynamic-resolution encoder \cite{bai2025qwen2}, one high-resolution webpage snapshot may require more than 10,000 tokens, resulting in a substantial computational cost and GPU memory allocation.

\begin{figure}[t]
    \centering
    \begin{minipage}{\linewidth}
        \centering
        \includegraphics[width=\linewidth]{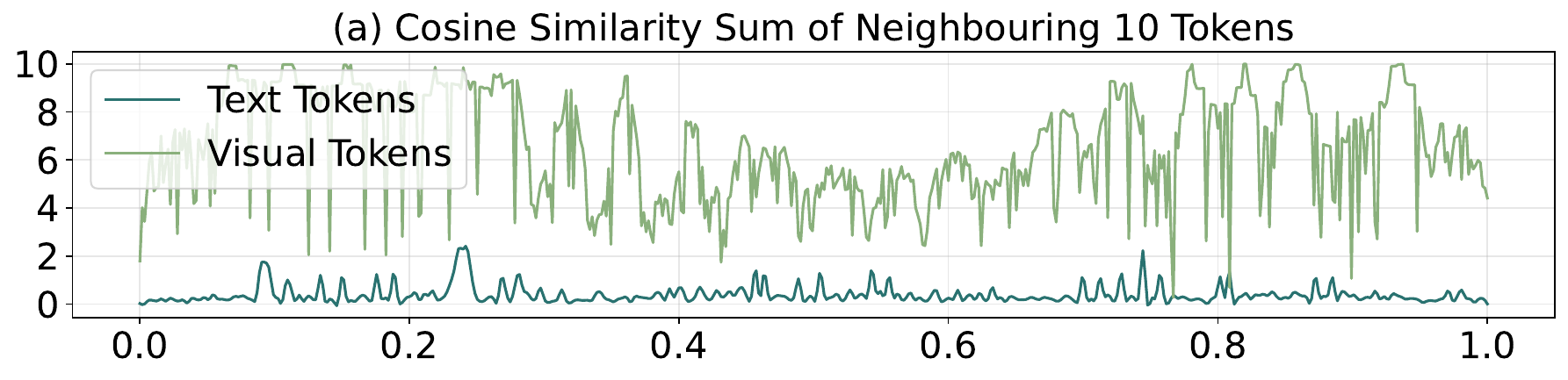}
    \end{minipage}
    \begin{minipage}{\linewidth}
        \centering
        \includegraphics[width=\linewidth]{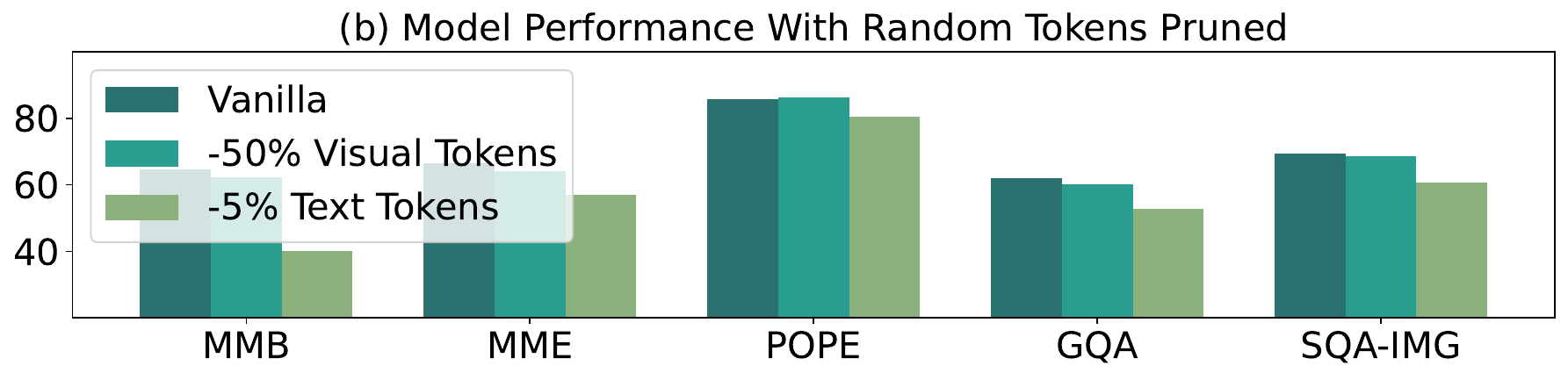}
    \end{minipage}
    \caption{\textbf{Redundancy analyses on visual tokens.} \textbf{(a)} The sum of cosine similarity between each token and neighbouring 10 tokens. \textbf{(b)} The performance of LLaVA-1.5-7B when randomly removing 50\% visual tokens or 5\% text tokens.}
    \label{fig:visualRedundancy}
\end{figure}

Although visual tokens constitute the majority of VLM input sequences, their necessity remains questionable. In Fig. \ref{fig:visualRedundancy}(a), we compare the cosine similarity between each token and its 10 neighbors and find that visual tokens exhibit significantly higher redundancy than text tokens. Fig. \ref{fig:visualRedundancy}(b) shows that randomly pruning 50\% of visual tokens causes a performance drop much smaller than removing merely 5\% of text tokens, while yielding substantial reductions in FLOPs. Moreover, previous works \cite{chen2024image, llavamini} observe that visual tokens receive markedly less attention in the LLM decoder compared to text tokens. These findings point to a key insight: visual tokens are highly redundant. Based on this, many works seek to reduce the number of visual tokens to overcome the computational burden. Some methods \cite{chen2024image, xing2024pyramiddrop, zhang2024sparsevlm} conduct token pruning inside the LLM decoder, while some \cite{yang2025visionzip, alvar2025divprune} conduct token selection based on static metrics like diversity or similarity. However, most methods do not fully utilize the intrinsic attention property of the vision encoder and are model-sensitive, necessitating careful tuning for practical deployment.

In this paper, we show that vision encoders process visual information in a progressive and structured hierarchy, where different layers attend to distinct semantic levels. Specifically, middle layers predominantly capture object-centric features, while deeper layers encode global contextual representations. This hierarchical pattern is consistently observed across various vision encoders \cite{radford2021learning, zhai2023sigmoid, touvron2021training, assran2025v}, regardless of the architecture design or pre-training data. Building on this observation, we introduce \textbf{HiPrune}, a training-free and model-agnostic visual token pruning plugin. We extract attention maps from a designated object layer $l$, selecting tokens with the highest attention scores and tokens near them as \textbf{Anchor Tokens} and \textbf{Buffer Tokens}, which encode detailed local semantics. The remaining token budget is allocated to \textbf{Register Tokens}, selected by the attention scores in the output layer, capturing global and holistic contextual features. We further introduce an optional subset of visual tokens selected based on similarity with text tokens, which showcases an improved instruction-following ability (HiPrune++). 

We evaluate our method across multiple popular VLMs. When applied to LLaVA-1.5, HiPrune maintains \textbf{99.3\%} of original performance while requiring only \textbf{1/3} tokens, alongside bringing a \textbf{58.7\%} FLOPs reduction. With a tight budget of \textbf{1/9} tokens, HiPrune++ still preserves \textbf{96.1\%} accuracy score, accompanied by an outstanding hallucination reduction compared with baselines. With a different encoder and dynamic token length setting, HiPurne achieves state-of-the-art on Qwen, confirming its versatility on various architectures.

Our main contributions are as follows:
\begin{itemize}
    \item We analyse the hierarchical attention patterns in vision encoders and reveal the focus of various layers qualitatively and quantitatively.
    \item We propose HiPrune and HiPrune++, a training-free and model-agnostic visual token pruning plugin enabling efficient inference.
    \item Extensive experiments on VLMs demonstrate the excellence of HiPrune, preserving \textbf{99.3\%} performance with only \textbf{1/3} visual tokens and reducing inference FLOPs by up to \textbf{58.7\%}.
\end{itemize}
\section{Related Works}
\subsection{Vision-Language Models}
VLMs \cite{bai2023qwen, wang-etal-2023-efficientvlm, chen2024far, zhang-etal-2025-sharper} have achieved impressive performance in various multimodal tasks. These models are typically composed of an image encoder, a projector, and an LLM. In the input sequence, visual tokens often constitute a significant portion. To improve the model's capacity for fine-grained understanding, some Vision-Language models, such as LLaVA-NeXT \cite{liu2024llavanext} and LLaVA-UHD \cite{guo2024llava}, increase the resolution of the input image, which further raises the number of visual tokens in the sequence. The excessive number of visual tokens leads to considerable computational overhead and adversely affects the inference speed of the model, constraining its practical deployment. This motivates the need for visual token compression techniques that reduce redundancy while preserving model performance.

\subsection{Visual Token Compression Methods}
Most visual token compression methods employ a pruning or merging strategy. FastV \cite{chen2024image} is a representative pruning method that compresses visual tokens by discarding those with low attention scores in the LLM. Following this work, some methods \cite{xing2024pyramiddrop, song2024less, hu-etal-2025-mplug} leverage text-image attention to prune tokens, which brings extra computational overhead. In transformer \cite{vaswani2017attention}, the attention score controls the information flow from layer to layer. Based on this theory, many approaches \cite{zhang2024cls, arif2025hired} overcome this by the attention from \texttt{CLS} token. However, not every vision encoder features such a token (e.g., SigLIP \cite{zhai2023sigmoid}), which constrains the adaptation of these methods for universal VLMs \cite{bai2025qwen2, chen2024far}. Apart from the drop strategy, merging-based approaches aim to reduce redundancy by fusing similar visual tokens. ToMe \cite{bolya2022token} performs token merging by aggregating visual tokens with high feature similarity. Subsequently, several studies \cite{cao-etal-2023-pumer, yang2025visionzip, huang-etal-2025-prunevid} have explored hybrid pruning-and-merging strategies. Most merging methods need extra training or functions, since merging is usually gradual across multiple layers and not compatible with FlashAttention \cite{dao2022flashattention}. In this work, we show that pruning tokens purely based on the vision encoder's inherent hierarchical attention pattern can achieve outstanding results without special tokens and unnecessary complexity.
\section{Motivated Insights}
\begin{figure}[t]
    \centering
    \begin{minipage}{0.49\linewidth}
        \centering
        \includegraphics[width=\linewidth]{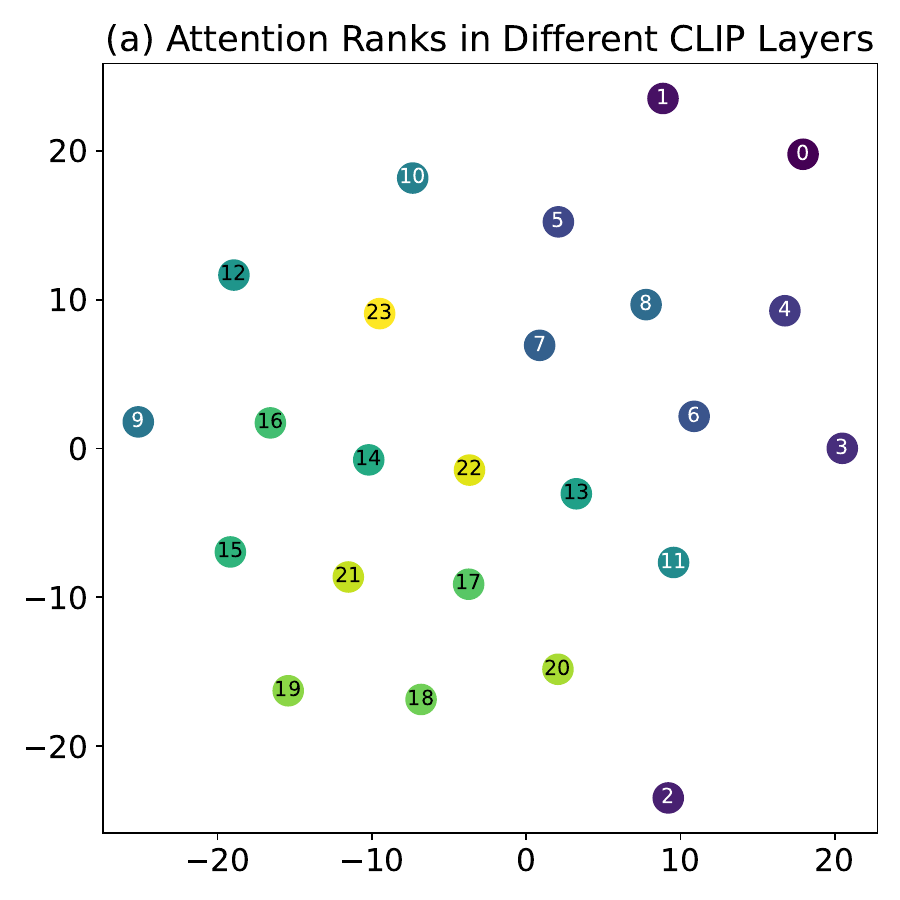}
    \end{minipage}
    \begin{minipage}{0.49\linewidth}
        \centering
        \includegraphics[width=\linewidth]{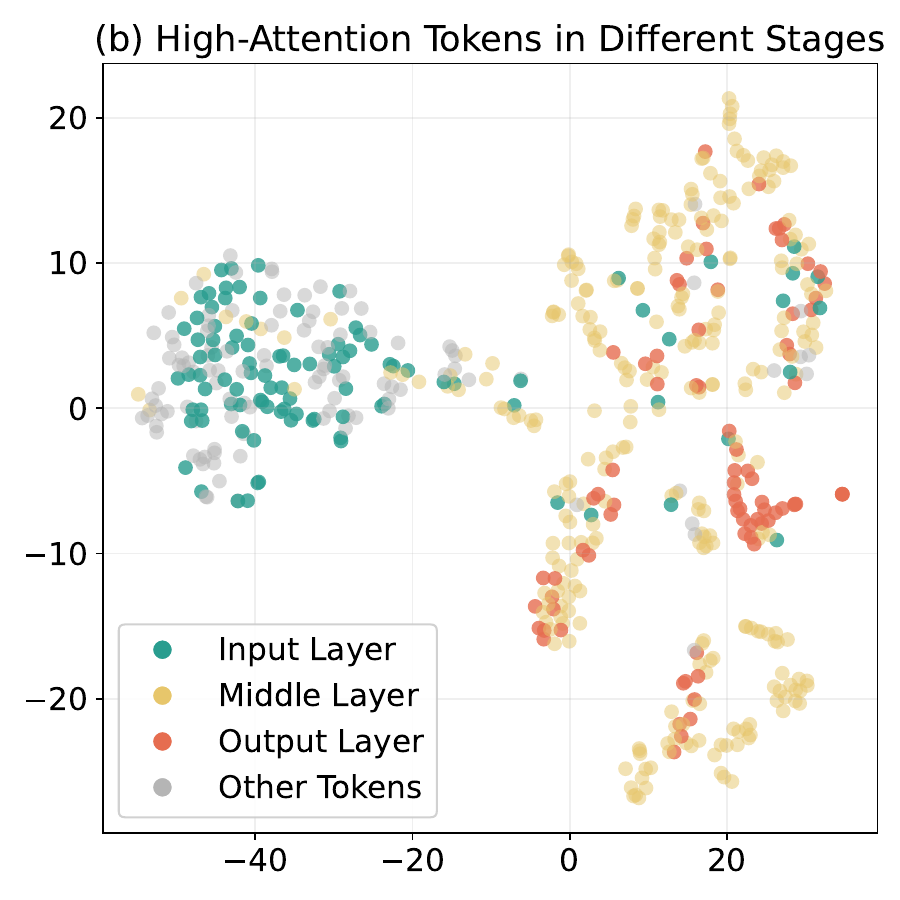}
    \end{minipage}
    \caption{\textbf{Hierarchical attention pattern in CLIP.} \textbf{(a)} Attention rankings of different layers. Most layers are adjacent to their neighbouring layers. \textbf{(b)} Top 50\% attentive tokens from different CLIP layers. The attention shifts from one cluster to another, showing a gradual and continuous transition, bridged by the middle-layer attention. Please refer to the Appendix \ref{sec:hie_attn_detail} for details.}
    \label{fig:attnTSNE}
\end{figure}

\begin{figure}[t]
    \centering
    \includegraphics[width=\linewidth]{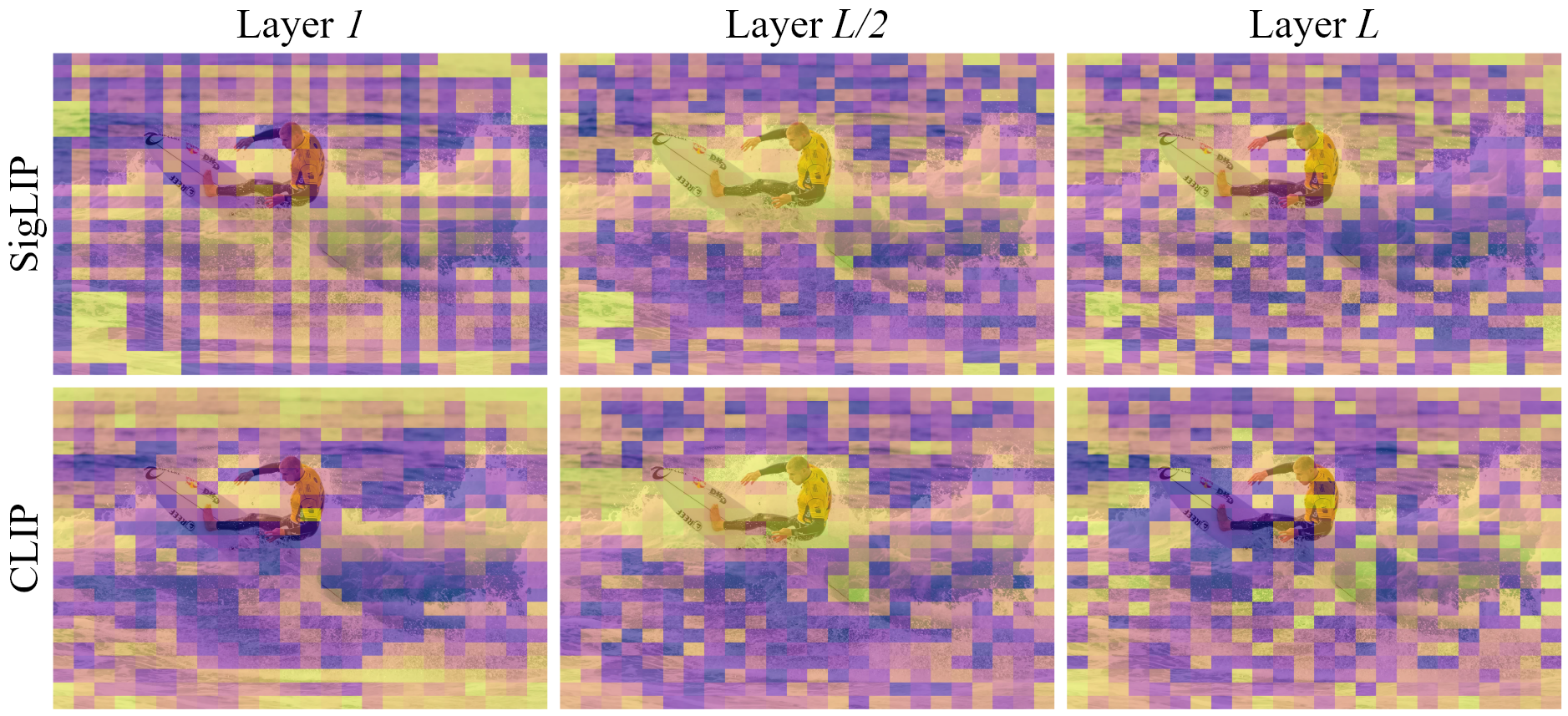}
    \caption{\textbf{Attention map for different layers of SigLIP and CLIP.} Patches with higher scores are in \textcolor{yellow}{\textbf{yellow}}. We can see that the middle layer is more object-centric.}
    \label{fig:attnVis}
\end{figure}

\begin{table}[t]
\centering
\resizebox{0.49\textwidth}{!}{
\small
\setlength{\tabcolsep}{1mm}
\begin{tabular}{cccccccc}
\toprule
 \textbf{Layer} & \textbf{CLIP-L} & \textbf{CLIP-B} & \textbf{SigLIP} & \textbf{SigLIP2} & \textbf{DeiT} & \textbf{VJEPA2} \\
 \midrule
1 & 0.58$\times$ & 0.34$\times$ & 0.57$\times$ & 0.62$\times$ & 0.27$\times$ & 0.82$\times$ \\
\textit{L/2} & \textbf{1$\times$} & \textbf{1$\times$} & \textbf{1$\times$} & \textbf{1$\times$} & \textbf{1$\times$} & \textbf{1$\times$} \\
\textit{L}  & 0.80$\times$ & 0.79$\times$ & 0.66$\times$ & 0.64$\times$ & 0.59$\times$ & 0.26$\times$ \\
\bottomrule
\end{tabular}
}
\caption{\textbf{IoU of object segmentation mask and top 10\% high-attention tokens.} Higher values stand for more overlap on objects in the image. `\textit{L}' denotes the total layers in the encoder. The data is normalized for a better comparison.}
\label{tab:Obj-IoU}
\end{table}

\paragraph{What Is the Representation Structure Inside Vision Encoders?}
To understand the focusing mechanism of vision encoders, we first take an insight into how attention distribution shifts across different layers. As shown in Fig. \ref{fig:attnTSNE}(a), we project the attention scores of different layers with t-SNE \cite{maaten2008visualizing}. The distribution reveals a continuous trajectory where most adjacent layers happen to be near each other, indicating the attention shifts across layers in a progressive and ordered way. We label tokens with high attention from different layers in Fig. \ref{fig:attnTSNE}(b). From the input to the output, the attention transfers from one cluster to another cluster, with the middle layer attention bridging both clusters. This continuous transition proves \textbf{the existence of an ordered representation hierarchy in the vision encoder}.

\paragraph{What do Middle Layers Focus on?}
We visualize the attention map for CLIP and SigLIP in Fig. \ref{fig:attnVis}. The attention distribution in the middle layers exhibits a distinct pattern: the model focuses on the main object of the image, e.g., the surf-man. To confirm this empirical observation, we compute the IoU between the object segmentation mask and top 10\% high-attention tokens in Table \ref{tab:Obj-IoU} using the COCO val2017 dataset \cite{lin2014microsoft}. The results show that the high-attention tokens from the middle layer share more overlap with objects than the input or output layer, indicating that \textbf{the attention from the middle layers is more correlated to objects in the image}. Notably, this tendency occurs across various encoders, including world models \cite{assran2025v}, showing little correlation with model architecture or training data.

\paragraph{What do Deep Layers Focus on?}
Previous works have argued that tokens receiving high attention scores in the deep layer of ViT encode rich global information by conducting image classification tasks on these tokens \cite{darcet2023vision}. In Fig. \ref{fig:attnVis}, we can see that in the output layer, the high-attention tokens diffuse across the whole image. Despite showing little correlation with the object, these tokens include patches of the image uniformly and can serve as an ideal indicator of the image under a limited token budget. Therefore, we can conclude that \textbf{tokens receiving high attention in the output layer encode global information}.

\section{Method}
An overview of our method is depicted in Fig. \ref{fig:method} and pseudo-code is given in Appendix \ref{sec:pseudo-code}. In the following, we first revisit the self-attention, then we present the design of HiPrune and HiPrune++.

\subsection{Preliminaries}
In a ViT-based vision encoder, an image is encoded into multiple tokens, forming a visual token matrix $\textbf{T}_v \in \mathbb{R}^{N \times d}$, where $N$ denotes the number of patches in an image and $d$ denotes the hidden dimension of the model. In each layer, the tokens are first mapped into $\textbf{Q},\textbf{K},\textbf{V} \in \mathbb{R}^{N \times d}$, and subsequently, the attention matrix $\textbf{A}$ is computed by
\begin{align}
    \textbf{A}=\operatorname{softmax}(\frac{\textbf{Q}\textbf{K}^T}{\sqrt{d_k}}) \in \mathbb{R}^{H \times N \times N}.
    \label{eq:attn}
\end{align}

The information sharing between tokens only takes place in Eq. \ref{eq:attn}. Intuitively, the more `important' a token is, the more its value in every token's new state, which is assigned by $\textbf{A}$. Therefore, in layer $l$, the importance of tokens can be weighted by their attention score:
\begin{align}
    \textbf{a}^{[l]}&=\frac{1}{H} \sum_{h=1}^H \sum_{n=1}^N \textbf{A}^{[l]}[h,n,:], \\
    &=(a_1^{[l]},a_2^{[l]},\dots,a_N^{[l]}) \in \mathbb{R}^N.
    \label{eq:score}
\end{align}

\subsection{Retained Tokens}
\paragraph{Anchor tokens} denote tokens with the highest attention score in the middle layers of the vision encoder. As discussed in our motivated insights, the middle layers tend to focus on object features, as evidenced by higher attention scores for tokens related to the object. Based on this, anchor tokens encode rich, detailed information about the raw image and should be retained when pruning. 

\paragraph{Buffer tokens} indicate tokens spatially adjacent to anchor tokens. As indicated by previous studies \cite{yang2024denoising}, noise exists in the attention map of ViTs. In Fig. \ref{fig:attnVis}, despite most high-attention tokens concentrating on the surf-man, a few tokens diffuse among the image, which may mislead the anchor tokens. To mitigate the noise issue and preserve spatial relationship, we include tokens neighbouring the anchor as a buffer.

\paragraph{Register tokens} receive top attention scores in the output layer of the vision encoder. In deep layers of the vision encoder, the high-attention tokens distribute uniformly across the image, serving as an ideal indicator of global information \cite{darcet2023vision}. To enhance the model's overall understanding of an image, we supplement the token set with register tokens, which is a common practice in approaches for VLM token pruning \cite{zhang2024cls, arif2025hired}.

\begin{figure}
    \centering
    \includegraphics[width=\linewidth]{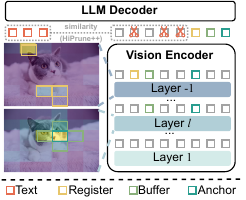}
    \caption{\textbf{Overview of HiPrune and HiPrune++.} HiPrune selects anchor and buffer tokens based on the attention from the object layer $l$, and register tokens from the last encoder layer. HiPrune++ additionally includes a small set of tokens selected by cosine similarity with text tokens to enhance instruction following ability.}
    \label{fig:method}
\end{figure}

\subsection{Pruning Pipeline}
Given a target token budget $N'$, we designate the object layer $l$ and object proportion $\alpha$, denoting the layer from which to determine anchor and buffer tokens, and the added-up proportion of them. Taking the \texttt{cross} strategy as an example (4 buffers around 1 anchor), we first draw the attention score $\mathbf{a}^{[l]}$ from the object layer $l$ and calculate the number of anchor tokens by $N_a=[\frac{\alpha\cdot N'}{5}]$. The anchor token indices set $\mathcal{I}_A$ is decided by
\begin{align}
    \mathcal{I}_A&=\{i\mid|\{j|a_j^{[l]}>a_i^{[l]}\}|<N_a\}.
\end{align}

Once the anchor tokens are decided, we proceed to retain buffer tokens. It is worth noting that the buffer selection scheme can be diverse, but little difference exists between choices as long as the size is big enough, which is to be discussed in our ablation studies. Assuming an $H_r\times W_c$ image is patchfied into $r\times c$ tokens, the buffer token indices $\mathcal{I}_B$ is calculated by
\begin{equation}
    \mathcal{I}_B=\cup \{\mathcal{I}_A-1,\mathcal{I}_A+1,\mathcal{I}_A-c,\mathcal{I}_A+c\} \cap [0,rc-1].
\end{equation}

The register token selection occurs after anchor tokens and buffer tokens, given the attention score $\mathbf{a}^{[-1]}$ of the output layer of the vision encoder, the registers are selected by
\begin{align}
    \mathcal{I}_R=\{i\mid |&\{j|a_j^{[-1]}>a_i^{[-1]}\}|<N'-|\mathcal{I}_A\cup \mathcal{I}_B| \\
    & \wedge i \notin \mathcal{I}_A\cup \mathcal{I}_B \}.
\end{align}

It is notable that $\mathcal{I}_A$, $\mathcal{I}_B$, and $\mathcal{I}_R$ are just the indices of tokens; the tokens for the LLM component are still chosen from the output layer of the vision encoder. After obtaining these token indices, HiPrune directly selects the corresponding tokens from the original token matrix $\textbf{T}_v \in \mathbb{R}^{N \times d}$ and discards the rest, leaving a pruned token matrix
\begin{equation}
    \mathbf{T'}_v=\mathbf{T}_v[\mathcal{I}_A \cup \mathcal{I}_B \cup \mathcal{I}_R,:]\in \mathbb{R}^{N' \times d}.
\end{equation}

\subsection{Text Guidance}

HiPrune is intentionally designed to be text-agnostic, meaning that the pruning process itself does not rely on any textual guidance. This design enables direct application to ViT-based vision models beyond the VLM paradigm, with more robustness and generalization. Nevertheless, HiPrune remains orthogonal to most text-aware token pruning methods and can be seamlessly combined with them. To confirm the compatibility, we introduce a text-aware extension \textbf{HiPrune++}, which performs lightweight text-relevance pruning after HiPrune.

Specifically, for each previously unselected visual token $\mathbf{t}_v^i \in \mathbb{R}^d, i \notin \mathcal{I}_A \cup \mathcal{I}_B \cup \mathcal{I}_R$, we compute its cosine similarity $\mathbf{r}^i$ with the averaged text embedding $\mathbf{t}_t \in \mathbb{R}^d$ by
\begin{equation}
    \mathbf{r}^i=\frac{\mathbf{t}_v^i \cdot \mathbf{t}_t}{\| \mathbf{t}_v^i \| \cdot \| \mathbf{t}_t \|}.
\end{equation}

Following the common practice \cite{zhang2025beyond}, for vision encoders with a paired text encoder \cite{radford2021learning}, we obtain $\mathbf{t}_t$ with the corresponding encoder. For those without a paired text encoder \cite{bai2025qwen2}, we use the average of all text embeddings. We then retain $[\beta \cdot N]$ visual tokens by the magnitude of $\mathbf{r}$, where $\beta$ is the proportion of visual tokens selected by text-relevance.

\section{Experiments}
\begin{table*}[t]
\centering
\setlength{\tabcolsep}{5pt}
\resizebox{\textwidth}{!}{
\begin{tabular}{ll|ccccccccc|c}
\toprule
\textbf{Method} & \textbf{\quad Venue\quad} & \textbf{GQA} & \textbf{MMB} & \textbf{MMB$^{\text{CN}}$} & \textbf{MME} & \textbf{POPE} & \textbf{SQA$^{\text{IMG}}$} & \textbf{VQA$^{\text{V2}}$} & \textbf{VQA$^\text{Text}$} & \textbf{VizWiz}  &{\textbf{Average}}\\ \midrule
\rowcolor{teal!20}
\multicolumn{12}{c}{\textit{Vanilla, 576 Tokens (100\%)}}\\ 
{LLaVA-1.5-7B} & {\textit{CVPR'24}} & {61.9} & {64.7} & {58.1} & {1862} & {85.9} & {69.5} & {78.5} & {58.2} & {50.0}  & {100.0\%} \\
\rowcolor{teal!10}
\multicolumn{12}{c}{\textit{Retain 192 Tokens (33.3 \%)}} \\
ToMe & \textit{ICLR'23} & 54.3 & 60.5 & - & 1563 & 72.4 & 65.2 & 68.0 & 52.1 & - &   88.5\% \\
FastV & \textit{ECCV'24} & 52.7 & 61.2 & {57.0} & 1612 & 64.8 & 67.3 & 67.1 & 52.5 & {50.8} &   90.4\% \\
HiRED$^\dagger$ & \textit{AAAI'25} & 58.8 & 62.6 & 54.5 & 1742 & 83.0 & 67.9 & 75.0 & - & 51.1 & 96.4\%\\
TRIM$^\dagger$ & \textit{COLING'25} &59.9 & 64.1 & 53.6 & 1765 & 87.1 & 67.8 & 76.2 &54.9  & 50.4 & 97.1\%\\
PyramidDrop & \textit{CVPR'25} & 57.3 & {63.3} & 56.8 & {1797} & 82.3 & {69.2} & 75.1 & 56.5 & {51.1} &  97.2\% \\
VisionZip & \textit{CVPR'25} & {59.3} & 63.0 & - & 1783 & {85.3} & 68.9 & {76.8} & {57.3} & - & 97.7\%   \\
SparseVLM$^\dagger$ & \textit{ICML'25} & 59.5 & 64.1 & 58.0 & 1780 & 85.4 & 68.8  & 77.0 & 57.7 & 50.6 & 98.6\%\\
\textbf{HiPrune} & \textbf{\textit{Ours}} & 59.2 & 62.8 & 57.0 & 1814 & 86.1 & 68.9 & 76.7 & 57.6 & 54.5 & \underline{99.3\%}\\
\textbf{HiPrune++} & \textbf{\textit{Ours}} & 60.3 & 63.5 & 57.5 & 1818 & 86.9 & 68.8 & 77.2 & 57.5 & 54.7 & \textbf{99.9\%} \\
\rowcolor{teal!10}
\multicolumn{12}{c}{\textit{Retain 128 Tokens (22.2\%)}} \\
ToMe & \textit{ICLR'23} & 52.4 & 53.3 & 48.8 & 1343 & 62.8 & 59.6 & 63.0 & 49.1 & 50.2 & 83.0\% \\
FastV & \textit{ECCV'24} & 49.6 & 56.1 & 56.4 & 1490 & 59.6 & 60.2 & 61.8 & 50.6 & 51.3 &   85.4\% \\
HiRED$^\dagger$ & \textit{AAAI'25} & 57.1 & 61.7 & 53.9 & 1714 & 79.8 & 68.1 & 73.5 & - & 51.4 & 95.0\%\\
TRIM$^\dagger$ & \textit{COLING'25} & 58.9 & 63.3 & 51.5 & 1732 & 87.2 & 68.4 & 74.8 & 52.7 & 50.6 & 95.7\%\\
PyramidDrop & \textit{CVPR'25} & 57.1 & 61.6 & {56.6} & 1761 & 82.3 & 68.4 & 72.9 & 56.6 & 51.0 &  {96.2\%} \\
VisionZip & \textit{CVPR'25} & {57.6} & {62.0} & - & {1762} & {83.2} & {68.9} & {75.6} & {56.8} & - & 96.2\% \\
SparseVLM$^\dagger$ & \textit{ICML'25} & 53.8 & 64.4 & 58.1 & 1761 &  85.0 & 68.5 & 76.3 & 56.7&  50.2& 97.0\% \\
\textbf{HiPrune} & \textbf{\textit{Ours}} & 57.3 & 62.2 & 56.4 & 1782 & 82.8 & 68.3 & 74.9 & 56.6 & 54.3 & \underline{97.5\%}\\
\textbf{HiPrune++} & \textbf{\textit{Ours}} & 59.0 & 62.3 & 57.0 & 1780 & 86.4 & 68.5 & 76.2 & 56.0 & 54.6 & \textbf{98.8\%} \\
\rowcolor{teal!10}
\multicolumn{12}{c}{\textit{Retain 64 Tokens (11.1\%)}} \\
ToMe & \textit{ICLR'23} & 48.6 & 43.7 & 38.9 & 1138 & 52.5 & 50.0 & 57.1 & 45.3 & 50.2 & 73.1\% \\
FastV & \textit{ECCV'24} & 46.1 & 48.0 & {52.7} & 1256 & 48.0 & 51.1 & 55.0 & 47.8 & {50.8} & 76.7\% \\
HiRED$^\dagger$ & \textit{AAAI'25} & 54.6 & 60.2 & 51.3 & 1595 & 73.7 & 68.2 & 69.8 & - & 53.3 & 91.8\% \\
TRIM$^\dagger$ & \textit{COLING'25} & 56.9& 61.5 & 44.9 & 1603 & 86.7 & 69.0& 71.9 & 50.0 & 50.6 & 92.1\%\\
PyramidDrop & \textit{CVPR'25} & 47.5 & 58.8 & 50.5 & 1561 & 55.9 & {69.0} & 69.2 & 50.6 & 50.7 &  86.6\% \\
VisionZip & \textit{CVPR'25} & {55.1} & {60.1} & - & {1690} & {77.0} & 69.0 & {72.4} & {55.5} & - & \underline{92.7\%} \\
SparseVLM$^\dagger$ & \textit{ICML'25} & 53.7 & 60.1 & 52.5 & 1559 & 77.5 & 69.7 & 70.2 &  53.4& 50.4 & 91.8\% \\
\textbf{HiPrune} & \textbf{\textit{Ours}} & 53.6 & 59.5 & 53.4 & 1646 & 73.0 & 68.9 & 69.2 & 54.9 & 54.4 & \underline{92.7\%}\\
\textbf{HiPrune++} & \textbf{\textit{Ours}} & 56.4 & 60.3 & 53.8 & 1767 & 84.3 & 68.9 & 72.8 & 54.5 & 54.7 & \textbf{96.1\%} \\
\bottomrule
\end{tabular}
}
\caption{\textbf{Results on LLaVA-1.5-7B.} `$^\dagger$' denotes our reproduced results, others are from \cite{zhang2025vscan}.}
\label{tab:llava}
\end{table*}

\subsection{Experiment Settings}
Following popular works \cite{zhang2024cls, chen2024image}, we conduct evaluations against other token pruning methods on four widely used VLMs, i.e., LLaVA-1.5-7B \cite{liu2024improved}, LLaVA-NeXT-7B \cite{liu2024llavanext}, Qwen2.5-VL-3B \cite{bai2025qwen2}, and Video-LLaVA \cite{lin2023video}. Model descriptions, benchmark datasets, and comparison details are in Appendix \ref{sec:eva_details}.

\paragraph{Implementation Details.}
We follow the default settings for each compared method as specified in their code repositories. In HiPrune, for both LLaVA-1.5-7B and LLaVA-NeXT-7B, we set $l=9$ and $\alpha=0.1$. For Qwen, we set $l=16$ and $\alpha=0.1$ since it has more layers in the vision encoder. We set $\beta=0.1$ for all the models when evaluating Hiprune++. Most of the evaluations are performed with the LMMs-Eval toolkit \cite{zhang2024lmmsevalrealitycheckevaluation}, and FLOPs are computed with the calflops package. All the experiments are conducted on one NVIDIA A100-PCIE (40G) unless otherwise specified.

\subsection{Accuracy Results}
\begin{table}[t]
    \centering
    \resizebox{0.48\textwidth}{!}{
    \setlength{\tabcolsep}{1mm}
    \begin{tabular}{lccccccc}
    \toprule
    \textbf{Method} & \textbf{MMB} & \textbf{MMB$^\text{CN}$} & \textbf{POPE} & \textbf{SQA$^\text{IMG}$} & \textbf{VizWiz} & \textbf{Avg} \\ \midrule
    \rowcolor{teal!20}
    \multicolumn{7}{c}{\textit{Vanilla, 2880 Tokens (100\%)}} \\
    LLaVA & 67.4 & 60.6 & 86.5 & 70.1 & 57.6 & 100\% \\ 
    \rowcolor{teal!10}
    \multicolumn{7}{c}{\textit{Retain 640 Tokens (22.2\%)}} \\
    HiRED & 66.0 & 57.0 & 85.0 & 68.3 & 59.1 & 98.1\%\\
    TRIM & 66.8 & 55.8 & 86.9 & 66.9 & 54.8 & 96.0\%\\
    VisionZip & 66.3 & 58.1 & 86.3 & 68.1 & 57.1 & 98.1\% \\
    DivPrune & 65.0 & 56.4 & 85.4 & 67.9 & 58.6 & 97.4\% \\
    VisPruner & 65.2 & 56.0 & 85.7 & 67.8 & 60.9 & 98.1\%\\
    \textbf{HiPrune} & 67.0 & 59.3 & 85.3 & 68.0 & 59.9 & \underline{99.4\%}\\
    \textbf{HiPrune++} & 67.2 & 59.1 & 87.1 & 67.8 & 59.9 & \textbf{99.7\%} \\
    \rowcolor{teal!10}
    \multicolumn{7}{c}{\textit{Retain 320 Tokens (11.1\%)}} \\
    HiRED & 64.2 & 56.4 & 83.3 & 66.8 & 58.3 & 96.2\%\\
    TRIM & 63.5 & 51.0 & 86.5 & 66.2 & 53.5 & 93.1\%\\
    VisionZip & 63.1 & 55.6 & 82.1 & 67.3 & 56.2 & 94.8\%\\
    DivPrune & 63.9 & 55.2 & 83.0 & 67.7 & 57.4 & 95.6\% \\
    VisPruner & 63.8 & 55.4 & 80.8 & 68.3 & 60.4 & \underline{96.4\%} \\
    \textbf{HiPrune} & 65.3 & 57.0 & 78.9 & 67.3 & 59.9 & \underline{96.4\%} \\
    \textbf{HiPrune++} & 66.2 & 57.4 & 85.6 & 67.2 & 60.1 & \textbf{98.4\%} \\
    \rowcolor{teal!10}
    \multicolumn{7}{c}{\textit{Retain 160 Tokens (5.6\%)}} \\
    TRIM & 61.6 & 45.2 & 84.8 & 65.5 & 52.9 & 89.9\%\\
    VisionZip & 60.1 & 50.4 & 74.8 & 68.3 & 55.5 & 90.5\%\\
    DivPrune & 62.5 & 52.3 & 78.4 & 68.3 & 57.5 & \underline{93.4\%} \\
    VisPruner & 59.2 & 51.3 & 73.5 & 68.9 & 60.1 & 92.0\% \\
    \textbf{HiPrune} & 59.8 & 50.7 & 67.7 & 68.7 & 57.2 & 89.6\%\\
    \textbf{HiPrune++} & 61.5 & 50.6 & 85.0 & 68.0 & 58.6 & \textbf{94.4\%} \\
    \bottomrule
    \end{tabular}
    }
    \caption{\textbf{Results on LLaVA-NeXT-7B.} The full table (Table \ref{tab:supp_llava-next}) for more comparisons is in Appendix \ref{sec:supp_exp}.}
    \label{tab:llava-next}
\end{table}
\begin{table}[t]
    \centering
    \resizebox{0.48\textwidth}{!}{
    \setlength{\tabcolsep}{1mm}
    \begin{tabular}{lccccccc}
    \toprule
    \textbf{Method} & \textbf{MMB} & \textbf{MMB$^\text{CN}$} & \textbf{POPE} & \textbf{SQA$^\text{IMG}$} & \textbf{VizWiz} & \textbf{Avg} \\ \midrule
    \rowcolor{teal!20}
    \multicolumn{7}{c}{\textit{Vanilla, 100\% Tokens}} \\
    Qwen & 77.3 & 73.0 & 87.0 & 80.4 & 68.3 & 100 \% \\ 
    \rowcolor{teal!10}
    \multicolumn{7}{c}{\textit{Retain 33.3\% Tokens}} \\
    FastV & 74.4 & 70.6 & 85.0 & 79.3 & 66.9 & 97.4\%\\ 
    VisionZip & 74.9 & 69.8 & 85.4 & 80.1 & 67.1 & 97.7\%\\ 
    \textbf{HiPrune} & 75.9 & 71.4 & 85.9 & 79.4 & 68.4 & \textbf{98.7\%}\\ 
    \textbf{HiPrune++} & 76.0 & 71.1 & 85.9 & 79.9 & 68.0 & \textbf{98.7\%} \\
    \rowcolor{teal!10}
    \multicolumn{7}{c}{\textit{Retain 22.2\% Tokens}} \\
    FastV & 72.4 & 69.2 & 82.7 & 79.6 & 66.2 & 95.9\%\\ 
    VisionZip & 73.5 & 67.4 & 84.6 & 80.0 & 66.3 & 96.2\%\\ 
    \textbf{HiPrune} & 73.7 & 69.1 & 84.9 & 80.2 & 67.1 & \textbf{97.1\%}\\ 
    \textbf{HiPrune++} & 74.2 & 69.4 & 84.4 & 80.2 & 66.7 & \textbf{97.1\%} \\
    \rowcolor{teal!10}
    \multicolumn{7}{c}{\textit{Retain 11.1\% Tokens}} \\
    FastV & 56.2 & 60.7 & 73.3 & 79.3 & 63.8 & 86.4\%\\ 
    VisionZip & 67.8 & 63.3 & 80.2 & 79.5 & 62.8 & 91.5\% \\ 
     \textbf{HiPrune} & 69.6 & 65.4 & 80.4 & 79.1 & 64.6 & \textbf{93.0\%}\\ 
     \textbf{HiPrune++} & 70.5 & 64.7 & 79.9 & 79.4 & 64.5 & \textbf{93.0\%} \\
    \bottomrule
    \end{tabular}
    }
    \caption{\textbf{Results on Qwen2.5-VL-3B-Instruct.} All the results are reproduced by us.}
    \label{tab:qwen}
\end{table}

\begin{table}[t]
\setlength{\tabcolsep}{1mm}
\centering
\resizebox{0.49\textwidth}{!}{
\begin{tabular}{lccccc}
\toprule
\textbf{Method} & \textbf{\begin{tabular}[c]{@{}c@{}}FLOPs\\ (T)$\downarrow$\end{tabular}} & \textbf{\begin{tabular}[c]{@{}c@{}}Prefill\\ (ms)$\downarrow$\end{tabular}} & \textbf{\begin{tabular}[c]{@{}c@{}}Decode\\ (ms)$\downarrow$\end{tabular}} & \textbf{\begin{tabular}[c]{@{}c@{}}Throughput\\ (tokens/s)$\uparrow$\end{tabular}} & \textbf{\begin{tabular}[c]{@{}c@{}}VRAM\\ (GB)$\downarrow$\end{tabular}} \\ \midrule
\rowcolor{teal!20}
\multicolumn{6}{c}{\textit{Vanilla, 576 Tokens (100\%)}} \\
\textcolor{gray}{LLaVA-7B} & \textcolor{gray}{$8.63$} & \textcolor{gray}{$54.31_{\pm0.37}$} & \textcolor{gray}{$21.85_{\pm0.26}$} & \textcolor{gray}{$44.42_{\pm0.37}$} & \textcolor{gray}{$14.52$} \\
\rowcolor{teal!10}
\multicolumn{6}{c}{\textit{Retain 192 Tokens (33.3\%)}} \\
HiPrune & $3.56$ & $28.56_{\pm0.30}$ & $21.65_{\pm0.08}$ & $45.65_{\pm0.07}$ & $14.52$ \\
HiPrune++ & $3.57$ & $28.89_{\pm0.11}$ & $21.73_{\pm0.15}$ & $45.23_{\pm0.08}$ & $15.38$ \\
\textcolor{gray}{Random} & \textcolor{gray}{$3.56$} & \textcolor{gray}{$28.83_{\pm0.14}$} & \textcolor{gray}{$20.97_{\pm0.07}$} & \textcolor{gray}{$47.17_{\pm0.05}$} & \textcolor{gray}{$14.52$} \\
\rowcolor{teal!10}
\multicolumn{6}{c}{\textit{Retain 128 Tokens (22.2\%)}} \\
HiPrune & $2.71$ & $25.59_{\pm0.17}$ & $21.53_{\pm0.08}$ & $45.59_{\pm0.07}$ & $14.35$ \\
HiPrune++ & $2.73$ & $25.76_{\pm0.16}$ & $21.59_{\pm0.10}$ & $45.47_{\pm0.07}$ & $15.38$ \\
\textcolor{gray}{Random} & \textcolor{gray}{$2.71$} & \textcolor{gray}{$25.52_{\pm0.14}$} & \textcolor{gray}{$20.59_{\pm0.15}$} & \textcolor{gray}{$47.59_{\pm0.20}$} & \textcolor{gray}{$14.07$} \\
\rowcolor{teal!10}
\multicolumn{6}{c}{\textit{Retain 64 Tokens (11.1\%)}} \\
HiPrune & $1.87$ & $21.96_{\pm0.10}$ & $21.33_{\pm0.10}$ & $45.93_{\pm0.07}$ & $14.35$ \\
HiPrune++ & $1.88$ & $22.03_{\pm0.11}$ & $21.41_{\pm0.12}$ & $45.80_{\pm0.10}$ & $15.38$ \\
\textcolor{gray}{Random} & \textcolor{gray}{$1.87$} & \textcolor{gray}{$21.81_{\pm0.15}$} & \textcolor{gray}{$21.00_{\pm0.39}$} & \textcolor{gray}{$46.80_{\pm0.64}$} & \textcolor{gray}{$14.02$} \\ \bottomrule
\end{tabular}
}
\caption{\textbf{Wall-clock latency and throughput.} `Random' randomly prunes tokens and adds no computational overhead, serving as a reference.}
\label{tab:efficiency}
\end{table}
\begin{table}[t]
\centering
\resizebox{0.49\textwidth}{!}{
\setlength{\tabcolsep}{1.7mm}
\begin{tabular}{lccccc}
\toprule
\textbf{Setting} & \textbf{GQA} & \textbf{MME} & \textbf{POPE} & \textbf{VizWiz} & \textbf{Avg} \\ \midrule
\rowcolor{teal!10}
\multicolumn{6}{c}{\textit{(a) Attention Pattern}} \\
\texttt{CLS} Token & 59.4 & 1772 & 85.4 & 55.5 & 99.8\%\\
Global* & 59.2 & 1814 & 86.1 & 54.5 & \textbf{100.0\%} \\
\rowcolor{teal!10}
\multicolumn{6}{c}{\textit{(b) Token Type}} \\
w/o Register & 58.4 & 1693 & 85.5 & 54.7 & 97.9\% \\
w/o Buffer & 59.1 & 1807 & 85.9 & 54.2 & 99.7\% \\
w/o Buf+Anc & 59.1 & 1805 & 85.9 & 54.4 & 99.7\% \\
Full* & 59.2 & 1814 & 86.1 & 54.5 & \textbf{100.0\%} \\
\rowcolor{teal!10}
\multicolumn{6}{c}{\textit{(c) Selection Scheme}} \\
Square(8) & 59.2 & 1817 & 86.0 & 54.4 & \textbf{100.0\%} \\
Row(2) & 59.2 & 1795 & 85.9 & 54.3 & 99.6\% \\
Cross(4)* & 59.2 & 1814 & 86.1 & 54.5 & \textbf{100.0\%} \\
\bottomrule
\end{tabular}
}
\caption{\textbf{Ablation study on HiPrune.} Each set is evaluated on LLaVA-1.5-7B with 192 tokens and $\alpha=0.1$. The number in row (c) denotes the buffer token number. `*' denotes the default setting.}
\label{tab:ablation}
\end{table}

\begin{figure}[t]
    \centering
    \includegraphics[width=\linewidth]{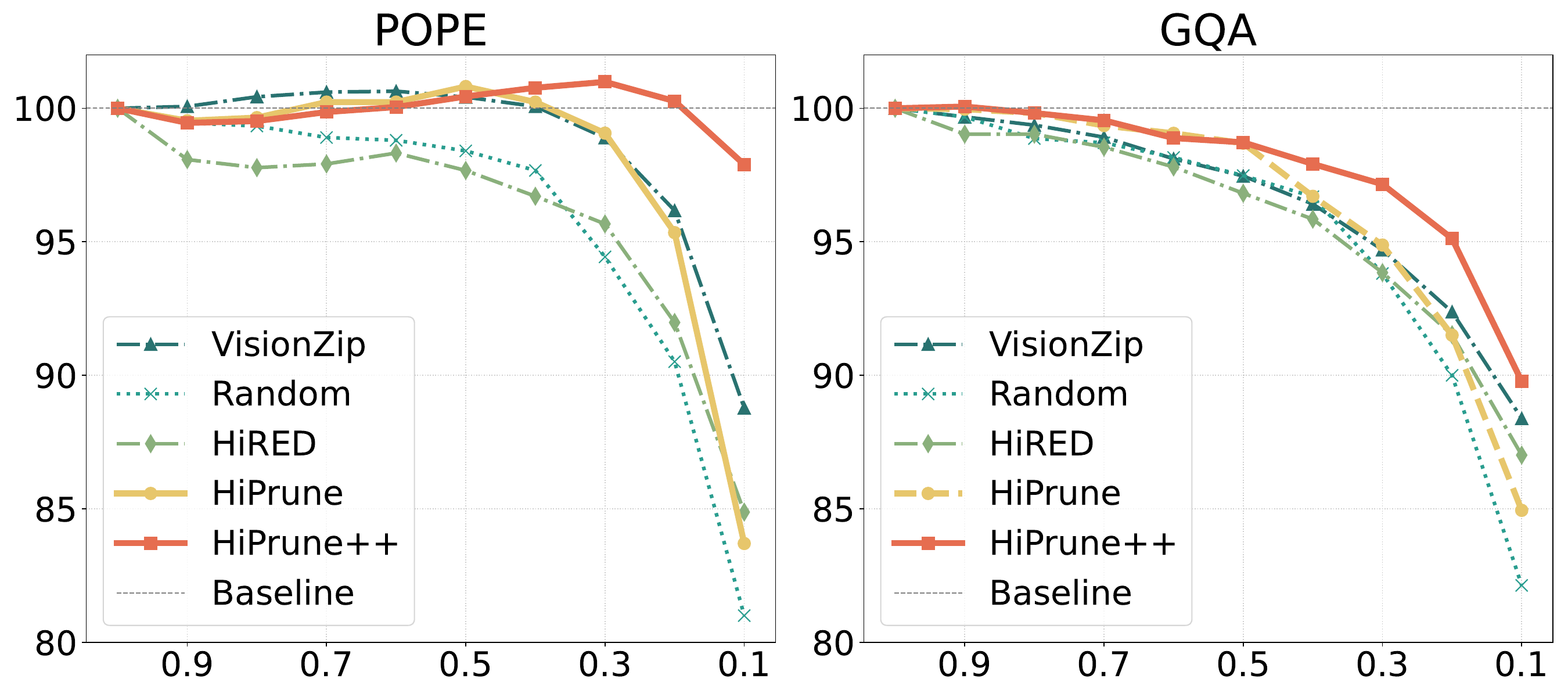}
    \caption{\textbf{Pareto analyses.} The horizontal axis is the token retention ratio, while the vertical axis is the percentage normalized accuracy results on LLaVA-1.5-7B.}
    \label{fig:pareto_on_llava}
\end{figure}

\paragraph{Results on LLaVA-1.5.} The accuracy results are shown in Table \ref{tab:llava} with 192, 128, and 64 tokens retained as the common practice. Across all the settings, HiPrune consistently outperforms existing methods, demonstrating superior performance. Specifically, with \textbf{1/3} tokens retained, HiPrune and HiPrune++ preserve \textbf{99.3\%} and \textbf{99.9\%} of the original model's average performance, almost matching the vanilla model. We present Pareto analyses on the token budget against hallucination and accuracy performance in Fig. \ref{fig:pareto_on_llava}. Interestingly, the superiority of HiPrune++ against HiPrune is more significant with fewer tokens, highlighting the necessity of text guidance under a lower budget.



\paragraph{Results on LLaVA-NeXT.} In Table \ref{tab:llava-next} we present evaluations on LLaVA-NeXT-7B \cite{liu2024llavanext}, a high-resolution VLM with more visual tokens. When retaining only \textbf{2/9} visual tokens, HiPrune++ preserves \textbf{99.7\%} accuracy, while HiPrune maintains \textbf{99.4\%}, which are quite close to the original model. With \textbf{11.1\%} and \textbf{5.6\%} of tokens retained, HiPrune++ still preserves \textbf{98.4\%} and \textbf{94.4\%} performance, respectively, demonstrating robustness on high-resolution models that handle more images and visual tokens.

\paragraph{Results on Qwen.} To verify the versatility of HiPrune, we further insert it into Qwen2.5-VL \cite{bai2025qwen2} in Table \ref{tab:qwen}. Unlike CLIP, some approaches relying on the text encoder or special tokens may have limited performance or cannot be implemented. When applied to Qwen, HiPrune achieves SOTA performance across the three settings. At an \textbf{11.1\%} retention rate, HiPrune preserves \textbf{93.0\%} of the model's original performance, outperforming FastV and VisionZip by \textbf{6.6\%} and \textbf{1.5\%}, respectively. The results on the Qwen series further support our key insights on vision encoders, regardless of their pre-training data or architecture.

\paragraph{Results on Video-LLaVA.} Video-LLaVA results are in Appendix \ref{sec:supp_exp} due to the space limitation.

\begin{figure}[t]
    \centering
    \includegraphics[width=\linewidth]{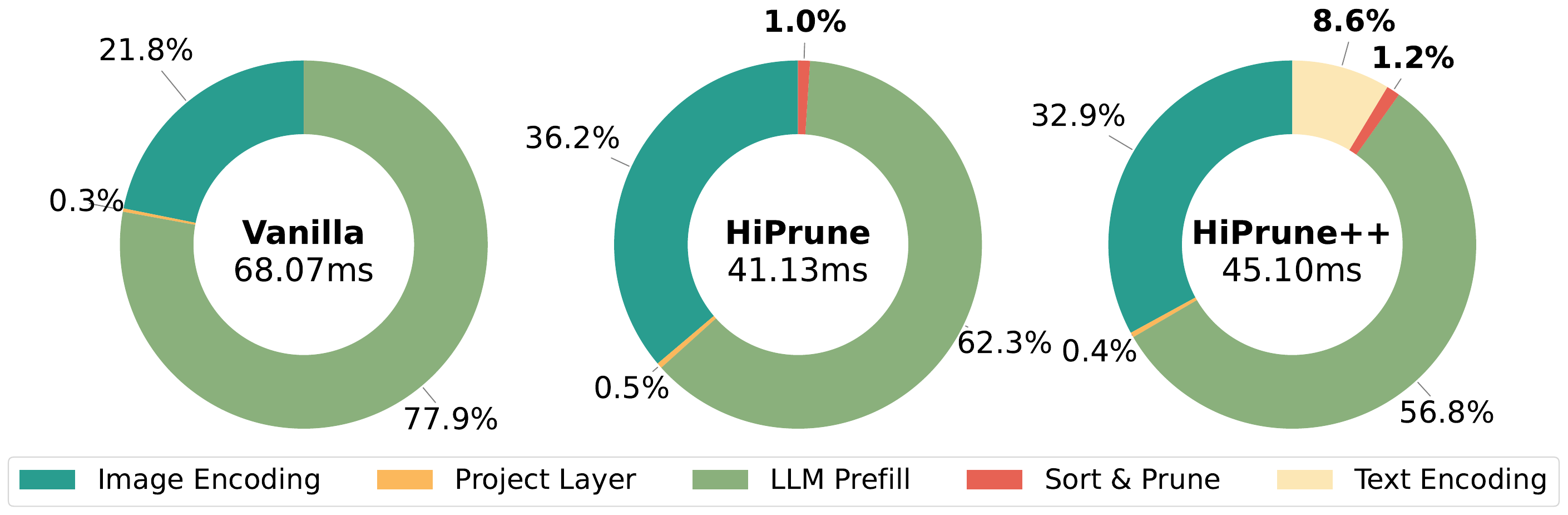}
    \caption{\textbf{Component overhead analyses.} The numbers inside each circle denote the wall-clock prefill latency for LLaVA-1.5-7B measured on RTX 5090. The budget for HiPrune and HiPrune++ is 192.}
    \label{fig:time_vis}
\end{figure}

\begin{figure*}[t]
    \centering
    \includegraphics[width=\linewidth]{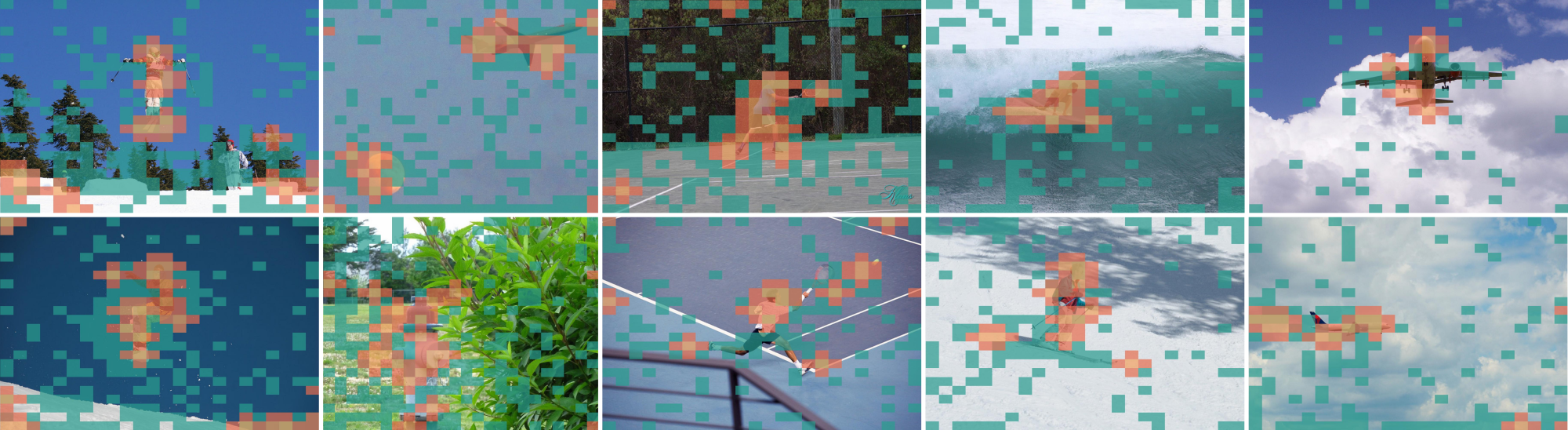}
    \caption{\textbf{Visualization on tokens retained by HiPrune.} Anchor tokens are in \textcolor{yellow}{\textbf{yellow}}, buffer tokens in \textcolor{orange}{\textbf{orange}}, and register tokens in \textcolor{teal}{\textbf{teal}}. Anchor and buffer tokens focus on the sports player, the aircraft, and the fire extinguisher. The images are slightly resized for better visualization and are randomly chosen from the COCO val2017 dataset.}
    \label{fig:ExpVis}
\end{figure*}

\subsection{Efficiency Results}
\label{sec:efficiency}
\paragraph{Latency and Throughput.}
We analyze the decoding latency and throughput of HiPrune on LLaVA with a simple example of around 600 tokens in Table \ref{tab:efficiency}. Compared with SparseVLM using 192 visual tokens, HiPrune achieves \textbf{1.32$\times$} faster during LLM prefill and gains a \textbf{1.20$\times$} speedup in generation, demonstrating its superior efficiency. When reducing the visual token number to 64, the FLOPs of a single forward pass are reduced by \textbf{78.3\%}, resulting in a \textbf{2.47$\times$} faster prefill.


\paragraph{Overhead Analyses.}

In Fig. \ref{fig:time_vis} we plot the time consumption of each component in HiPrune and HiPrune++ during the prefill phase. Both methods operate before the LLM decoder and thus are compatible with FlashAttention\cite{dao2022flashattention}. The sort introduced by HiPrune consumes only 1\% of the total time, which is negligible compared to the largely reduced prefill latency. With a text encoder, HiPrune++ still maintains a total overhead of under 10\%, which is acceptable in deployment.


\subsection{Ablation Studies}
We provide ablation studies on key designs in this subsection. Extended studies on hyperparameters and buffer selections are provided in Appendix \ref{sec:extended_ablation}.

\paragraph{Attention Pattern.} Many methods prune tokens guided by the \texttt{CLS} token's attention to other tokens; however, not every model has this token, making these methods model-specific. We compare computing attention score $\textbf{a}^{[l]}$ by Eq. \ref{eq:score} and \texttt{CLS} in Table \ref{tab:ablation}(a). For our method, the global attention achieves slightly better results and features much stronger versatility since it is model-agnostic.

\paragraph{Token Types.} We examine the retained anchor, buffer, and register tokens in HiPrune. As Table \ref{tab:ablation}(b) shows, removing either type degrades the model's performance. Specifically, removing register tokens causes the most significant degradation, highlighting the critical role of global information, which is in line with previous studies \cite{yang2025visionzip, zhang2024cls}.

\paragraph{Buffer Selection Scheme.} In Table \ref{tab:ablation}(c) we present different buffer token selection schemes. A square or cross can both achieve similar results; however, when the size of the buffer is too small or missing (row 2 in (b)), the results begin to drop. The buffer tokens are introduced to resist the noise in the attention map. When the window is too small, the effects of the buffer become limited.

\begin{figure}[t]
    \centering
    \includegraphics[width=\linewidth]{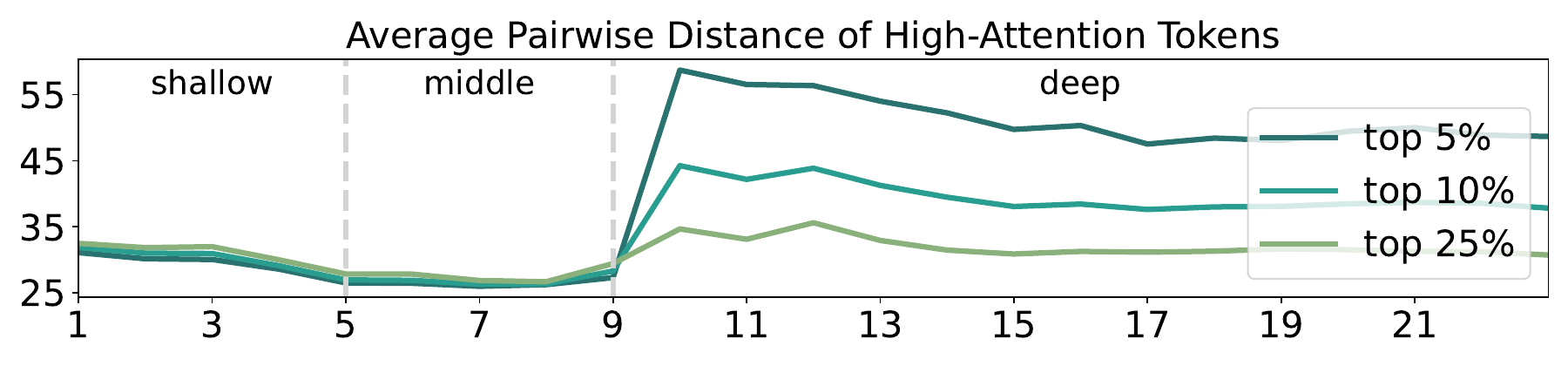}
    \includegraphics[width=\linewidth]{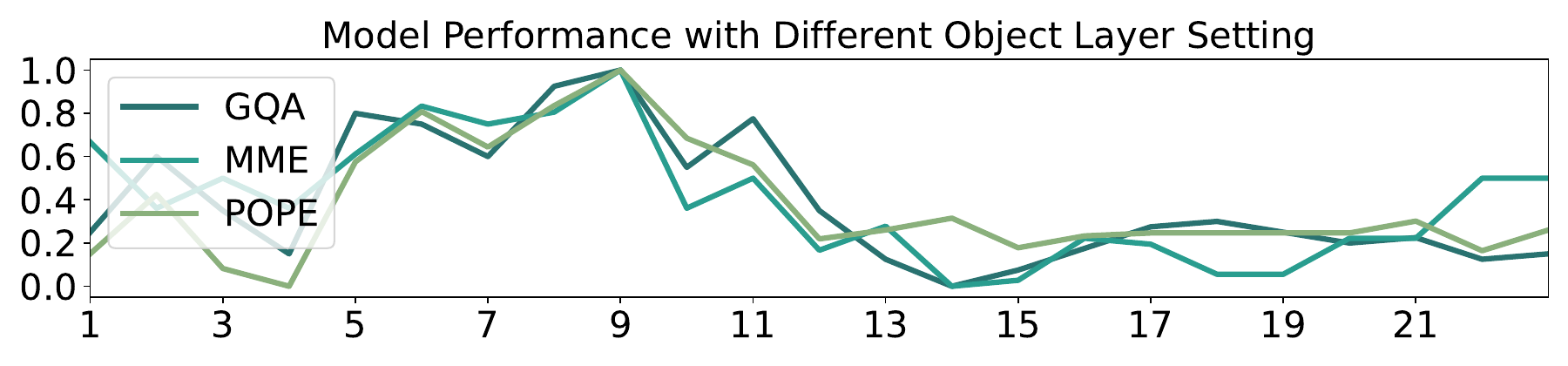}
    \caption{\textbf{Average pairwise distance of high-attention tokens and model performance with different Object Layer.} Based on this trend, we divide CLIP into three phases and set the Object Layer at the critical point.}
    \label{fig:layer_dist}
\end{figure}

\paragraph{Object Layer Setting.} We adopt a dispersion-based searching strategy to decide the Object Layer $l$, where the attention score for object and buffer tokens is extracted. Intuitively, tokens on the same object should feature high similarity. We plot the average pairwise distance of high-attention tokens and model performance with different $l$ in Fig. \ref{fig:layer_dist}. For LLaVA, model performance achieves an optimal when setting $l$ as 9. We assume that at this layer, tokens with similar semantic information are close in the embedding space, and at a critical point from object-centric to global information.


\subsection{Visualizations}
\paragraph{Selected Tokens.} In Fig. \ref{fig:ExpVis} we present a visualization of retained tokens. Anchor tokens and buffer tokens are mainly distributed on the main objects of the image, such as the body of the player, the aircraft in the sky, etc. Preserving these tokens can bring more information about image details and alleviate hallucination. The register tokens diffuse among the whole image uniformly. Despite showing little correlation with image segmentation, they carry indispensable global information. The reason has been discussed in our Motivated Insights. The combination of these tokens strikes a balance between overall and detail information. More visualizations are in Appendix \ref{sec:supp_vis}.


\section{Conclusion}
In this paper, we investigate the layer-wise attention patterns of vision encoders and reveal that middle layers predominantly capture object-centric features, while deeper layers emphasize global representations. Motivated by this, we propose HiPrune, a model-agnostic and training-free token pruning method that leverages the hierarchical attention structure within the vision encoder. HiPrune mainly preserves anchor, buffer, and register tokens, and an optional set of visual tokens selected by similarity with text embeddings. Extensive experiments across diverse VLMs demonstrate the robustness and generality of HiPrune, which achieves state-of-the-art results while significantly reducing computational overhead. We believe our findings offer valuable perspectives on the internal representation of vision encoders, and HiPrune will facilitate more efficient deployment of VLMs and inspire future research in this direction.

\section*{Acknowledgement}
This work is supported in part by the National Natural Science Foundation of China under grants U24A20328, 62301189, 62476071, 62576122, 62536003, 62521006, Guangdong Basic and Applied Basic Research Foundation under grants 2025A1515011732, 2026A1515011139.

\section*{Limitations}
HiPrune is most readily applicable to open-source models where architectural modifications are feasible, and cannot be incorporated into commercial models like Gemini and ChatGPT due to their black-box nature. Additionally, HiPrune relies on the multi-head self-attention in the transformer architecture and is not directly transferable to models using a CNN encoder. Besides, we only provide an explicit and simple way to couple HiPrune and text guidance. Despite the outstanding results, future work may focus on taking a better approach to couple text guidance and HiPrune.

\bibliography{custom}

\appendix
\newpage

\section{Pseudo-code for HiPrune}
\label{sec:pseudo-code}
In Algorithm \ref{alg:HiPrune}, we provide a pseudo-code for HiPrune and HiPrune++ written in PyTorch style \cite{paszke2019pytorch} to better explain our method. This example is adapted from LLaVA and adopts the \texttt{cross} as the strategy to select buffer tokens.

\begin{algorithm}[h]
\caption{HiPrune and HiPrune++}
\label{alg:HiPrune}
\textbf{Input}: Image tensor \texttt{image} \\
\textbf{Parameter}: Token Budget \texttt{N}, Object Layer \texttt{l}, Object Proportation \texttt{alpha}, Encoder Patch Size \texttt{p} \\
\textbf{Output}: Pruned token tensor \texttt{retained\_tokens}

\definecolor{codebg}{rgb}{0.95,0.95,0.97}
\definecolor{codeblue}{rgb}{0.25,0.5,0.5}
\definecolor{codegreen}{rgb}{0.1,0.5,0.2}
\definecolor{codegray}{rgb}{0.5,0.5,0.5}

\begin{lstlisting}[
    language=Python,
    basicstyle=\ttfamily\fontsize{6.7pt}{6.7pt}\selectfont,
    backgroundcolor=\color{codebg},
    keywordstyle=\color{blue},
    commentstyle=\color{codegreen},
    % numberstyle=\tiny\color{codegray},
    numbers=left,
    numbersep=6pt,
    frame=single,
    rulecolor=\color{white},
    breaklines=true,
    tabsize=2,
    xleftmargin=10pt,
    xrightmargin=10pt,
    showstringspaces=false
]
image_tokens, all_attns = encoder(image)
## Compute attention score from object layer l
mid_attn = all_attns[l].squeeze(0) # Remove batch
mid_attn = mid_attn.mean(0) # Average multi-head
mid_attn = mid_attn.sum(0) # Attention to each token
mid_attn = mid_attn[1:] # Exclude cls
## Assign anchor tokens
a_sum = round(N * alpha / 5) # 5 tokens in a cluster
a_idx = topk(mid_attn, k=a_sum).indices
## Assign buffer tokens
b_idx = cat([a_idx-1, a_idx+1, a_idx-p, a_idx+p])
a_b_idx = unique(cat([a_idx, b_idx]).clamp(0, p**2))
## Compute attention score from output layer
deep_attn = all_attns[-1].squeeze(0)
deep_attn = mid_attn.mean(0).sum(0)[1:]
## Assign register tokens
mask = zeros(N).scatter_(a_b_idx, 1)
deep_attn -= mask # Mask already-chosens
r_sum = N - a_b_idx.shape[0]
r_idx = topk(deep_attn, k=r_sum).indices
## Text Guidance in HiPrune++
t_sum = round(N * beta  / 5)
text_tokens = text_encoder(text)
avg_text_tokens = text_tokens.mean(-2)
avg_text_tokens /= avg_text_tokens.norm(-1)
image_tokens /= image_tokens.norm(-1)
similarity = avg_text_tokens @ image_tokens
mask = mask.scatter_(r_idx, 1)
similarity -= mask # Mask already-chosens
t_idx = topk(similarity, k=t_sum)
## Retain these tokens
retained_idx = cat([a_idx, b_idx, r_idx, t_idx])
retained_tokens = image_tokens[retained_idx]
return retained_tokens
\end{lstlisting}
\end{algorithm}

\section{Hierarchical Attention Pattern Details}
\label{sec:hie_attn_detail}
In Fig. \ref{fig:attnTSNE}, we show how attention distribution shifts across layers in CLIP. Here, we explain details about Fig. \ref{fig:attnTSNE}(a) and Fig. \ref{fig:attnTSNE}(b). Since HiPrune prunes visual tokens by their rankings rather than absolute values, we focus on the ranking of each token's attention. In Alg. \ref{alg:Fig2_a}, we state our acquisition process of Fig. \ref{fig:attnTSNE}(a), which shows the ranking of attentions across layers. In Fig. \ref{fig:attnTSNE}(b), it is worth noting that each dot (regardless of color) is the projected token \textbf{from the output layer}, and the color does not mean that the token is drawn from middle layers. We have included a comprehensive visualization in the next section.

\begin{algorithm}[t]
\textbf{Input}: Image tensor \texttt{image} \\
\textbf{Output}: Attention Ranking Coordinates \texttt{coor\_2d}
\caption{Acquisition process of Fig. \ref{fig:attnTSNE}(a).}
\label{alg:Fig2_a}

\definecolor{codebg}{rgb}{0.95,0.95,0.97}
\definecolor{codeblue}{rgb}{0.25,0.5,0.5}
\definecolor{codegreen}{rgb}{0.1,0.5,0.2}
\definecolor{codegray}{rgb}{0.5,0.5,0.5}

\begin{lstlisting}[
    language=Python,
    basicstyle=\ttfamily\fontsize{6.7pt}{6.7pt}\selectfont,
    backgroundcolor=\color{codebg},
    keywordstyle=\color{blue},
    commentstyle=\color{codegreen},
    % numberstyle=\tiny\color{codegray},
    numbers=left,
    numbersep=6pt,
    frame=single,
    rulecolor=\color{white},
    breaklines=true,
    tabsize=2,
    xleftmargin=10pt,
    xrightmargin=10pt,
    showstringspaces=false
]
all_ranks = []
tokens, all_attns = encoder(image)
for attn in all_attns:
    ## Same attn extraction process in Alg. 1
    attn = attn.squeeze(0)
    attn = attn.mean(0).sum(0)[1:]
    ## Transform attn values into ranks
    ## [0.2, 0.3, 0.5, 0.1] -> [1, 2, 3, 0]
    attn_rank = argsort(argsort(attn)) # [1, 576]
    all_ranks.append(attn_rank)
all_ranks = stack(all_ranks, dim=0) # [24, 576]
tsne = TSNE(n_components=2) # Project into 2D space
coor_2d = tsne.fit_transform(all_ranks) # [24, 2]
return coor_2d # plt.scatter(coor_2d)
\end{lstlisting}
\end{algorithm}

\begin{table*}[t]
\centering
\renewcommand{\arraystretch}{1.2}
\setlength{\tabcolsep}{5pt}
\resizebox{\textwidth}{!}{
\small
\begin{tabular}{lc|ccccccccc}
\toprule
\textbf{Method} & \textbf{Token Num} & \textbf{GQA} & \textbf{MMB} & \textbf{MMB}$^{\text{CN}}$ & \textbf{MME} & \textbf{POPE} & \textbf{SQA}$^{\text{IMG}}$ & \textbf{VQA}$^{\text{V2}}$ & \textbf{VQA$^\text{Text}$} & \textbf{VizWiz}\\
\midrule
\rowcolor{teal!20}
LLaVA-1.5-13B & 576 & 63.2 & 67.7 & 63.5 & 1818 & 85.9 & 72.8 & 80.0 & 61.3 & 53.6 \\
\multirow{3}{*}{w/ HiPrune} & 192 & 59.4 & 67.1 & 62.5 & 1798 & 85.4 & 73.7 & 78.0 & 59.5 & 55.6  \\
 & 128 & 57.9 & 66.8 & 63.1 & 1730 & 82.8 & 74.1 & 76.1 & 58.3 & 54.9 \\
 & 64 & 54.2 & 64.8 & 59.2 & 1634 & 72.4 & 74.6 & 70.3 & 56.7 & 56.0 \\ \midrule
 \multirow{3}{*}{w/ HiPrune++} & 192 & 60.2 & 66.5 & 62.5 & 1808 & 86.7 & 73.2 & 78.5 & 59.5 & 55.4\\
 & 128 & 59.1 & 67.2 & 62.8 & 1745 & 86.2 & 73.8 &77.4 &58.8 & 55.4\\
 & 64 & 56.9 & 65.0 &58.3 & 1736 & 84.4 & 74.1 & 73.9 &56.7 & 56.0\\ \midrule
\rowcolor{teal!20}
 LLaVA-NeXT-13B & 2880 & 64.4 & 68.5 & 61.2 & 1901 & 85.3 & 73.1 & 82.3 & 63.2 & 59.1 \\
\multirow{3}{*}{w/ HiPrune} & 640 & 62.6 & 70.2 & 65.3 & 1877 & 84.9 & 71.6 & 80.0 & 61.6 & 61.1 \\
 & 320 & 59.3 & 68.6 & 64.9 & 1800 & 79.7 & 72.1 & 75.7 & 60.1 & 59.2 \\
 & 160 & 54.4 & 66.3 & 61.3 & 1647 & 71.0 & 70.9 & 68.4 & 57.1 & 55.9 \\ \midrule
\multirow{3}{*}{w/ HiPrune++} & 640 & 63.5 & 69.9 & 64.8 & 1894 & 86.6 & 72.2 & 80.5 & 61.5 & 60.7 \\
 & 320 & 61.5 & 68.8 & 63.4 & 1823 & 86.3 & 71.2 & 77.6 & 59.8 & 59.1 \\
 & 160 & 58.2 & 66.3 & 60.7 & 1774 & 86.2 & 71.2 & 72.3 & 55.9 & 56.7 \\
\bottomrule
\end{tabular}
}
\caption{Performance comparisons on LLaVA-1.5-13B and LLaVA-NeXT-13B \cite{liu2024improved}.}
\label{tab:supp_llava}
\end{table*}

\begin{table*}[t]
\centering
\renewcommand{\arraystretch}{1.2}
\setlength{\tabcolsep}{5pt}
\small
\begin{tabular}{lc|cccccccc}
\toprule
\textbf{Method} & \textbf{Token Budget} & \textbf{GQA} & \textbf{MMB} & \textbf{MMB}$^{\text{CN}}$ & \textbf{MME} & \textbf{POPE} & \textbf{SQA}$^{\text{IMG}}$ & \textbf{VQA$^\text{text}$} & \textbf{VizWiz}  \\
\midrule
\rowcolor{teal!20}
Qwen2.5-VL-3B & 100\% & 59.9 & 77.3 & 73.0 & 2144 & 87.0 & 80.4 & 77.8 & 68.9 \\
\multirow{3}{*}{w/ HiPrune} & 33.3\% & 57.5 & 75.9 & 71.8 & 2061 & 86.0 & 79.8 & 70.1 & 68.0 \\
 & 22.2\% & 55.6 & 73.7 & 69.1 & 2002 & 84.5 & 80.0 & 62.9 & 67.0\\
 & 11.1\% & 51.5 & 69.7 & 65.1 & 1881 & 80.0 & 79.4 & 50.9 & 64.5 \\ \midrule
 \multirow{3}{*}{w/ HiPrune++} & 33.3\% & 57.4 & 75.6 & 70.9 & 2069 & 85.9 & 79.7 & 69.2 & 68.2 \\
 & 22.2\% & 55.7 & 73.5 & 68.9 & 1980 & 84.7 & 79.8 & 61.7 & 66.9 \\
 & 11.1\% & 51.3 & 69.8 & 64.9 & 1844 & 80.0 & 79.1 & 48.8 & 64.4 \\ \midrule
\rowcolor{teal!20}
Qwen2.5-VL-7B & 100\% & 60.5 & 83.2 & 80.1 & 2331 & 86.2 & 87.4 & 83.1 & 70.4 \\
\multirow{3}{*}{w/ HiPrune} & 33.3\% & 58.9 & 82.6 & 79.5 & 2297 & 85.1 & 87.0 & 78.5 & 69.2 \\
 & 22.2\% & 57.2 & 80.2 & 77.6 & 2168 & 84.0 & 85.7 & 74.1 & 68.6 \\
 & 11.1\% & 52.5 & 76.1 & 73.5 & 1998 & 80.2 & 82.8 & 62.6 & 66.4 \\ \midrule
 \multirow{3}{*}{w/ HiPrune++} & 33.3\% & 58.6 & 82.1 & 79.2 & 2310 & 85.1 & 86.6 & 77.9 & 69.0 \\
 & 22.2\% & 57.3 & 80.3 & 77.0 & 2172 & 83.6 & 85.2 & 73.2 & 68.5 \\
 & 11.1\% & 52.8 & 75.6 & 73.3 & 1999 & 79.5 & 82.8 & 60.7 & 66.7 \\
\bottomrule
\end{tabular}
\caption{Performance comparisons on Qwen2.5-VL-3B-Instruct and Qwen2.5-VL-7B-Instruct \cite{bai2025qwen2}.}
\label{tab:supp_qwen}
\end{table*}

\begin{table}[t]
    \centering
    \resizebox{0.48\textwidth}{!}{
    \setlength{\tabcolsep}{1mm}
    \begin{tabular}{lccccccc}
    \toprule
    \textbf{Method} & \textbf{MMB} & \textbf{MMB$^\text{CN}$} & \textbf{POPE} & \textbf{SQA$^\text{IMG}$} & \textbf{VizWiz} & \textbf{Avg} \\ \midrule
    \rowcolor{teal!20}
    \multicolumn{7}{c}{Vanilla, 2880 Tokens (100\%)} \\
    LLaVA & 67.4 & 60.6 & 86.5 & 70.1 & 57.6 & 100\% \\ 
    \rowcolor{teal!10}
    \multicolumn{7}{c}{Retain 640 Tokens (22.2\%)} \\
    FastV & 63.1 & 53.5 & 79.5 & 67.4 & 53.9 & 92.7\%\\
    HiRED & 66.0 & 57.0 & 85.0 & 68.3 & 59.1 & 98.1\%\\
    TRIM & 66.8 & 55.8 & 86.9 & 66.9 & 54.8 & 96.0\%\\
    VisionZip & 66.3 & 58.1 & 86.3 & 68.1 & 57.1 & 98.1\% \\
    DivPrune & 65.0 & 56.4 & 85.4 & 67.9 & 58.6 & 97.4\% \\
    PDrop & 64.1 & 55.2 & 83.8 & 66.7 & 53.8 & 94.3\%\\
    VisPruner & 65.2 & 56.0 & 85.7 & 67.8 & 60.9 & 98.1\%\\
    SparseVLM & 65.9 & 58.6 & 85.3 & 67.6 & 53.6 & 96.5\%\\
    \textbf{HiPrune} & 67.0 & 59.3 & 85.3 & 68.0 & 59.9 & {99.4\%}\\
    \textbf{HiPrune++} & 67.2 & 59.1 & 87.1 & 67.8 & 59.9 & \textbf{99.7\%} \\
    \rowcolor{teal!10}
    \multicolumn{7}{c}{Retain 320 Tokens (11.1\%)} \\
    FastV & 53.4 & 42.5 & 49.5 & 66.6 & 51.3 & 78.1\%\\
    HiRED & 64.2 & 56.4 & 83.3 & 66.8 & 58.3 & 96.2\%\\
    TRIM & 63.5 & 51.0 & 86.5 & 66.2 & 53.5 & 93.1\%\\
    VisionZip & 63.1 & 55.6 & 82.1 & 67.3 & 56.2 & 94.8\%\\
    DivPrune & 63.9 & 55.2 & 83.0 & 67.7 & 57.4 & 95.6\% \\
    PDrop & 55.5 & 44.7 & 60.8 & 66.7 & 49.7 & 81.6\%\\
    VisPruner & 63.8 & 55.4 & 80.8 & 68.3 & 60.4 & 96.4\% \\
    SparseVLM & 63.1 & 56.7 & 76.9 & 67.2 & 54.2 & 93.2\%\\
    \textbf{HiPrune} & 65.3 & 57.0 & 78.9 & 67.3 & 59.9 & {96.4\%}\\
    \textbf{HiPrune++} & 66.2 & 57.4 & 85.6 & 67.2 & 60.1 & \textbf{98.4\%} \\
    \rowcolor{teal!10}
    \multicolumn{7}{c}{Retain 160 Tokens (5.6\%)} \\
    TRIM & 61.6 & 45.2 & 84.8 & 65.5 & 52.9 & 89.9\%\\
    VisionZip & 60.1 & 50.4 & 74.8 & 68.3 & 55.5 & 90.5\%\\
    DivPrune & 62.5 & 52.3 & 78.4 & 68.3 & 57.5 & {93.4\%} \\
    VisPruner & 59.2 & 51.3 & 73.5 & 68.9 & 60.1 & 92.0\% \\
    \textbf{HiPrune} & 59.8 & 50.7 & 67.7 & 68.7 & 57.2 & 89.6\%\\
    \textbf{HiPrune++} & 61.5 & 50.6 & 85.0 & 68.0 & 58.6 & \textbf{94.4\%} \\
    \bottomrule
    \end{tabular}
    }
    \caption{\textbf{Results on LLaVA-NeXT-7B.} Comparison results are reported from \cite{zhang2025beyond}.}
    \label{tab:supp_llava-next}
\end{table}
\begin{table}[t]
    \centering
    \resizebox{0.48\textwidth}{!}{
    \setlength{\tabcolsep}{1mm}
    \begin{tabular}{lccccccc}
    \toprule
    \textbf{Method} & \textbf{MMB} & \textbf{MMB$^\text{CN}$} & \textbf{POPE} & \textbf{SQA$^\text{IMG}$} & \textbf{VizWiz} & \textbf{Avg} \\ \midrule
    \rowcolor{teal!20}
    \multicolumn{7}{c}{Vanilla, 100\% Tokens} \\
    Qwen & 86.3 & 84.0 & 84.2 & 91.5 & 64.8 & 100\% \\ 
    \rowcolor{teal!10}
    \multicolumn{7}{c}{Retain 22.2\% Tokens} \\
    FastV & 76.4 & 72.9 & 64.9 & 84.4 & 63.5 & 88.6\%\\ 
    VisionZip & 82.2 & 79.1 & 79.2 & 86.5 & 64.7 & 95.6\%\\ 
    \textbf{HiPrune} & 82.6 & 80.0 & 80.8 & 87.0 & 64.2 & \textbf{96.2\%}\\ 
    \textbf{HiPrune++} & 82.6 & 79.0 & 80.1 & 86.5 & 64.2 & 95.7\%\\
    \rowcolor{teal!10}
    \multicolumn{7}{c}{Retain 11.1\% Tokens} \\
    FastV & 67.2 & 63.5 & 41.0 & 82.9 & 61.9 & 77.7\%\\ 
    VisionZip & 76.0 & 73.9 & 71.7 & 83.5 & 62.6 & 89.8\%\\ 
    \textbf{HiPrune} & 77.2 & 74.3 & 71.7 & 82.6 & 63.0 & \textbf{90.1\%}\\ 
    \textbf{HiPrune++} & 76.6 & 73.5 & 72.1 & 82.9 &62.5 & 89.8\%\\
    \rowcolor{teal!10}
    \multicolumn{7}{c}{Retain 5.6\% Tokens} \\
    FastV & 56.4 & 53.0 & 34.3 & 78.7 & 60.1 & 69.6\% \\ 
    VisionZip & 69.8 & 66.6 & 62.5 & 81.5 & 60.9 & 83.5\%\\ 
    \textbf{HiPrune} & 71.3 & 67.4 & 63.8 & 82.6 & 61.1 & \textbf{84.7\%}\\ 
    \textbf{HiPrune++} & 70.3 & 67.4 & 63.3 & 81.8 & 60.8 & 84.0\% \\
    \bottomrule
    \end{tabular}
    }
    \caption{\textbf{Results on Qwen2.5-VL-32B-Instruct.} All the results are reproduced by us.}
    \label{tab:qwen_32B}
\end{table}
\begin{table}[t]
\renewcommand{\arraystretch}{1.2}
\centering
\resizebox{0.49\textwidth}{!}{
\setlength{\tabcolsep}{1.7mm}
\begin{tabular}{lccccc}
\toprule
\textbf{Setting} & \textbf{GQA} & \textbf{MME} & \textbf{POPE} & \textbf{VizWiz} & \textbf{Avg} \\ \midrule
Square(8) & 59.2 & 1817 & 86.0 & 54.4 & \textbf{100.0\%} \\
Cross(4)* & 59.2 & 1814 & 86.1 & 54.5 & \textbf{100.0\%} \\
Rot-Cross(4) & 59.3 & 1819 & 85.7 & 54.4 & \textbf{100.0\%} \\
Row(2) & 59.2 & 1795 & 85.9 & 54.3 & 99.6\% \\
Column(2) & 59.1 & 1805 & 85.7 & 54.3 & 99.7\% \\
\bottomrule
\end{tabular}
}
\caption{\textbf{Study on selection schemes of buffer tokens.} Each set is evaluated on LLaVA-1.5-7B with 192 tokens and $\alpha=0.1$. The number in denotes buffer token number. `*' denotes the default setting.}
\label{tab:buffer_selection}
\end{table}

\section{Evaluation Details}
\label{sec:eva_details}
\paragraph{Models.}
HiPrune is model-agnostic and training-free, applicable to any VLM with at least one vision encoder and an LLM. We conduct HiPrune on models with various vision encoders and visual token partition strategies. Following most previous work, we evaluate on LLaVA-1.5-7B \cite{liu2024improved} and LLaVA-NeXT-7B \cite{liu2024llavanext}, which encode images into fixed-length token sequences. We also include evaluations on Qwen2.5-VL-7B-Instruct and Qwen2.5-VL-32B-Instruct \cite{bai2025qwen2}, which utilizes a dynamic-resolution ViT and encodes images into sequences of varying lengths. For video evaluations, we apply HiPrune on Video-LLaVA \cite{lin2023video}. All the models utilized in this paper are downloaded from Huggingface. 

\paragraph{Comparisons.}
We compare HiPrune with 9 SOTA visual token reduction methods: ToMe \cite{bolya2022token}, FastV \cite{chen2024image}, SparseVLM \cite{zhang2024sparsevlm}, HiRED \cite{arif2025hired}, TRIM \cite{song2024less}, VisionZip \cite{yang2025visionzip}, and PyramidDrop \cite{xing2024pyramiddrop}. Some comparisons on Qwen are missing because the corresponding method either can only be applied to LLaVA or does not open-source code.

Among these, ToMe employed a fusion strategy, while FastV accelerated the inference process by reducing unnecessary tokens. SparseVLM utilized sparsity technology to compress tokens in the language model. HiRED decreased model complexity by selectively retaining important tokens. TRIM optimized processing speed and memory usage by eliminating unnecessary tokens. PyramidDrop applied a pyramid structure to reduce tokens layer by layer, and VisionZip enhanced efficiency by intelligently selecting and compressing tokens.

\paragraph{Datasets.}
We conduct thorough experiments across various multimodal benchmarks, including visual question answering benchmarks such as GQA \cite{hudson2019gqa}, SQA \cite{lu2022learn}, VQAv2 \cite{balanced_vqa_v2}, MME \cite{fu2024mmecomprehensiveevaluationbenchmark}, and TextVQA \cite{singh2019towards}. Additionally, we include POPE \cite{li2023evaluating} and VizWiz \cite{gurari2018vizwiz} to study the hallucination when visual tokens are pruned. We also include MMB and MMB-CN \cite{liu2024mmbench} to study the multilingual ability of VLM since some approaches rely on the CLIP text encoder to work and draw back on non-English tasks. For video tasks, we adopt MVBench \cite{li2024mvbench} and Vinoground \cite{zhang2024vinoground} to evaluate the model's overall performance in multiple domains.

\paragraph{Toolkits.} For HiPrune, most of our evaluations are completed with LMMs-Eval toolkit \cite{zhang2024lmmsevalrealitycheckevaluation}. However, since some benchmarks either are extremely slow on LMMs-Eval or need Internet for an online evaluation, the MMB, MMB-CN \cite{liu2024mmbench}, TextVQA \cite{singh2019towards}, and VQAv2 \cite{balanced_vqa_v2} results in Table \ref{tab:llava} are obtained with the public codebase released by LLaVA \cite{liu2023visual}. The rest of the results in Table \ref{tab:llava} and all the results in Tables \ref{tab:llava-next} and \ref{tab:qwen} are obtained with LMMs-Eval.

\begin{figure}[t]
    \centering
    \includegraphics[width=0.6\linewidth]{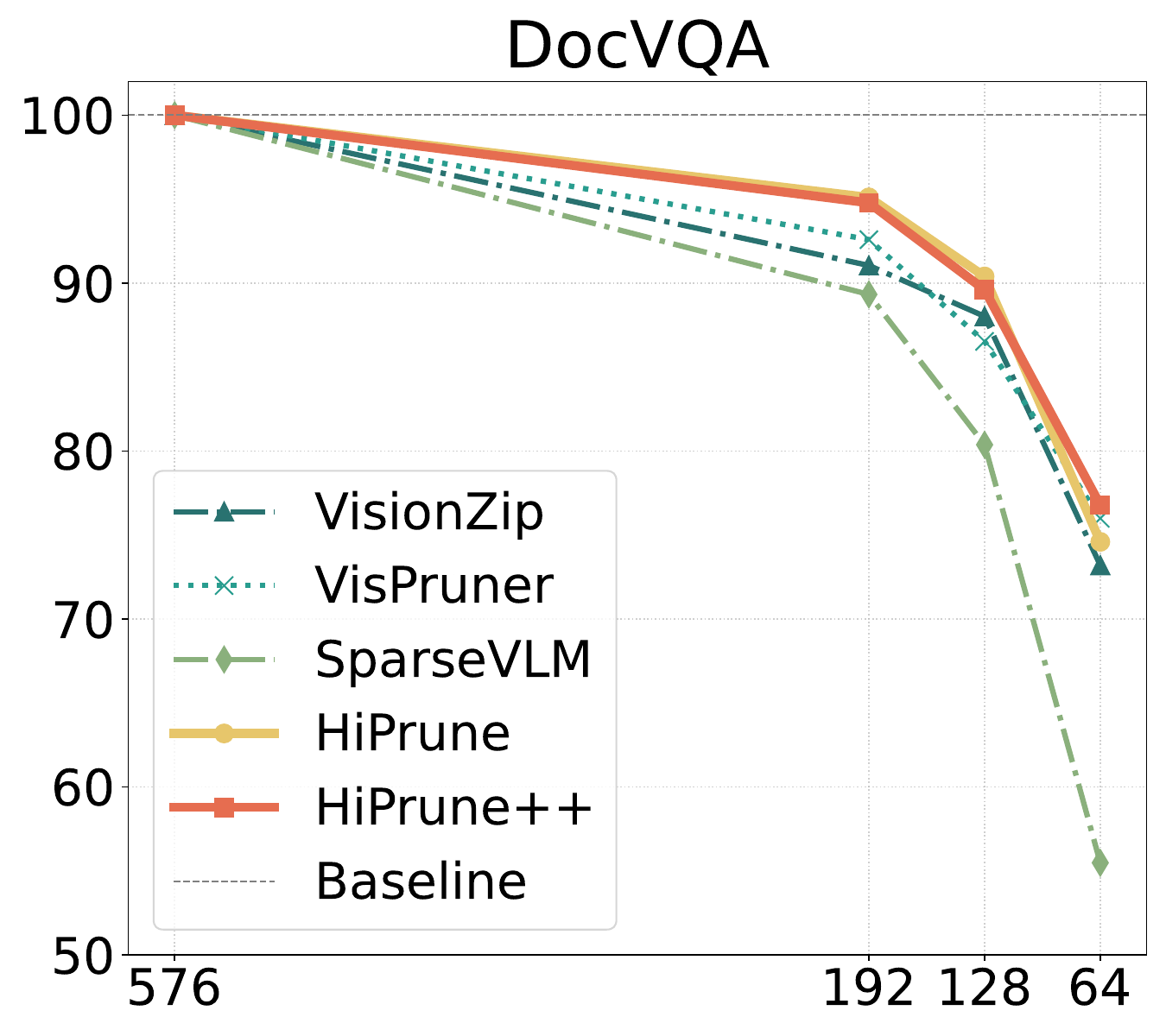}
    \caption{\textbf{Results on Text-Dominant Tasks.} We evaluate HiPrune on LLaVA-1.5-7B with the DocVQA dataset. The horizontal axis is the token budget, while the vertical axis is the percentage normalized results.}
    \label{fig:docvqa}
\end{figure}

\begin{figure}[t]
    \centering
    \includegraphics[width=\linewidth]{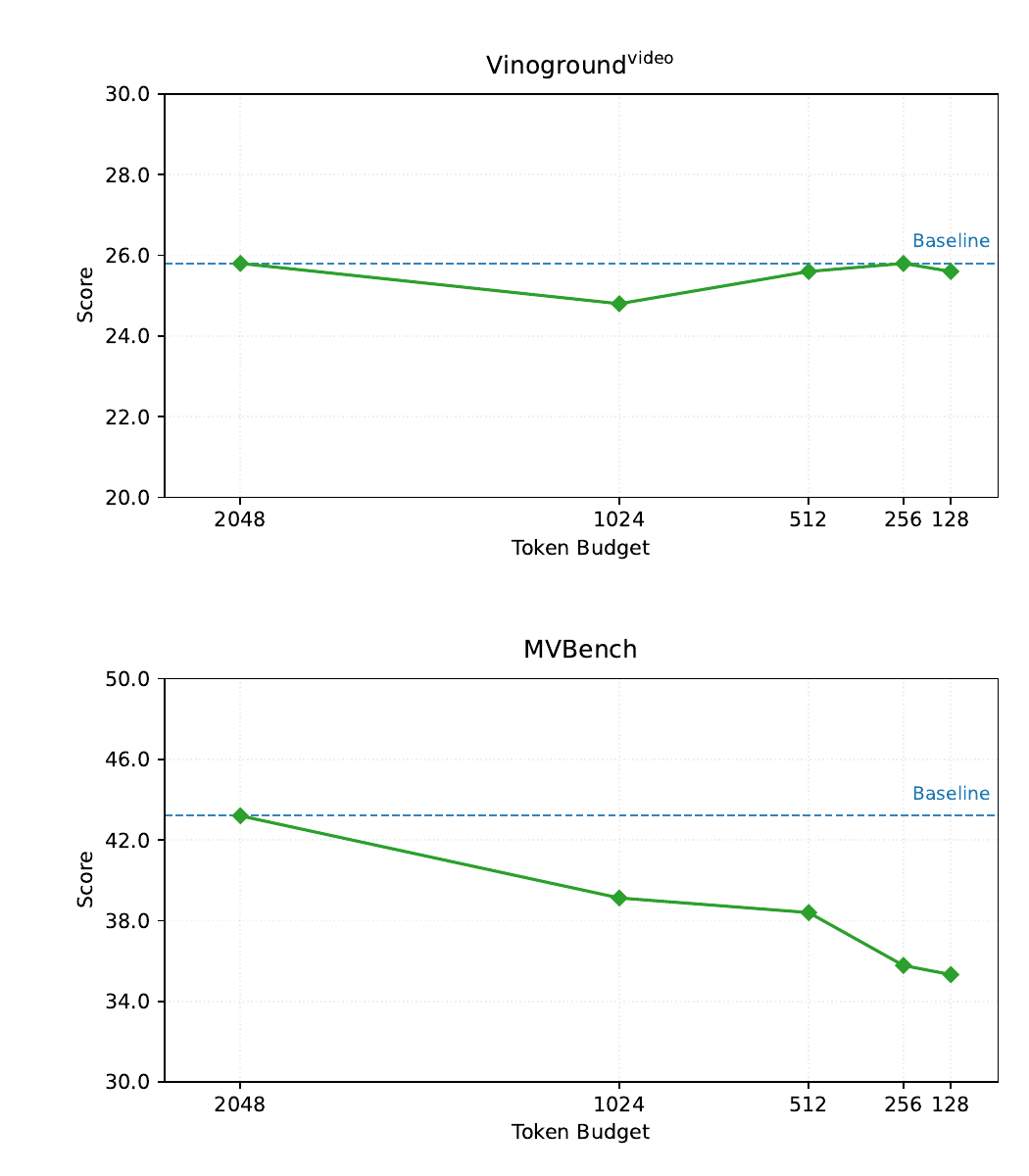}
    \caption{\textbf{Results on Video-LLaVA-7B.} We apply HiPrune and set different token budgets. The vanilla model features a token budget of 2048.}
    \label{fig:video-llava}
\end{figure}

\begin{figure}[t]
    \centering
    \includegraphics[width=\linewidth]{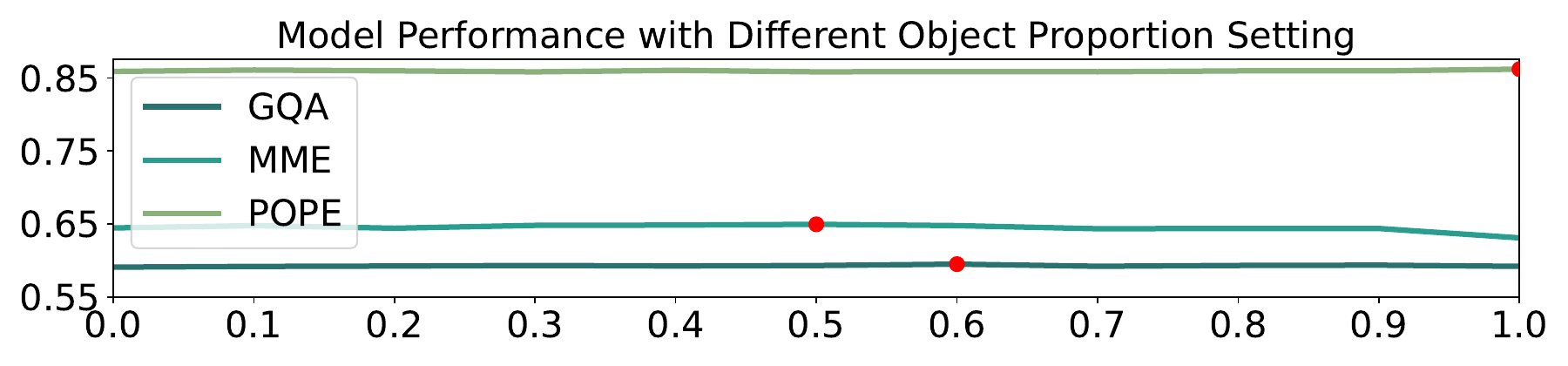} \hfill
    \includegraphics[width=\linewidth]{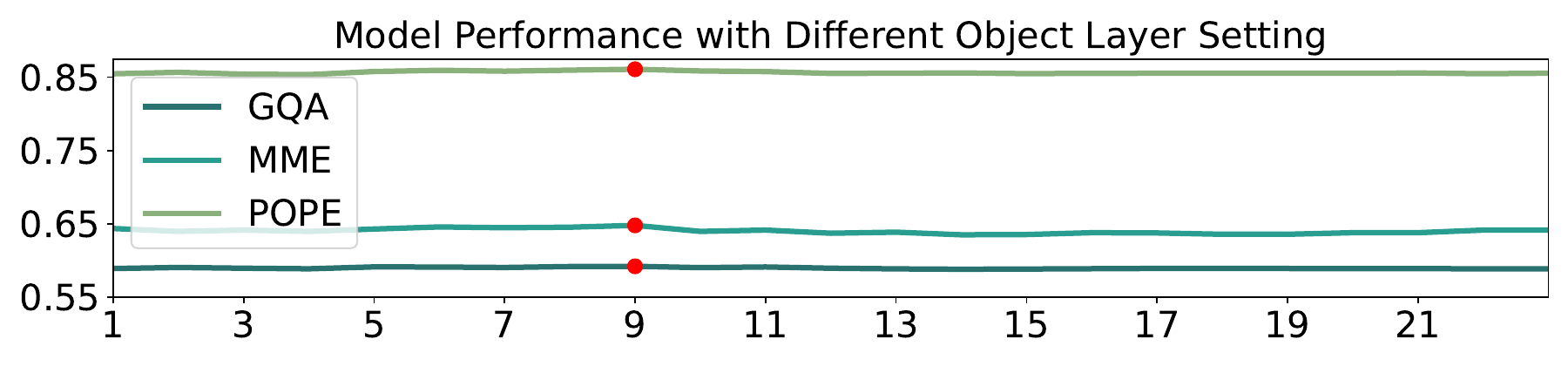}
    \caption{\textbf{Hyperparameter sensitivity.} The results are obtained on LLaVA-v1.5-7B with budget $N'=192$. The best performance setting is marked in \textcolor{red}{\textbf{red}}.}
    \label{fig:hyperparam}
\end{figure}

\begin{figure}[t]
    \centering
    \includegraphics[width=0.9\linewidth]{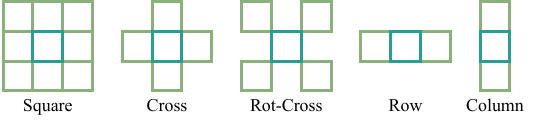}
    \caption{Positional relation between buffer and anchor tokens under different selection schemes. The anchor tokens are in \textcolor{teal}{\textbf{teal}} while the buffer tokens are in \textcolor{lime}{\textbf{lime}}.}
    \label{fig:buffer_selection}
\end{figure}

\section{Extended Experiments}
\subsection{Accuracy Results}
\label{sec:supp_exp}
\paragraph{Video Evaluations.} 
We apply HiPrune on Video-LLaVA-7B \cite{lin2023video}. As Fig. \ref{fig:video-llava} shows, for Vinoground, the accuracy results remain stable and keep \textbf{99.2\%} performance with \textbf{1/16} visual tokens. For MVBench, the performance drop is slightly observable but still acceptable.

\paragraph{Different Model Sizes.}
We further apply HiPurne to LLaVA-1.5-13B, LLaVA-NeXT-13B, Qwen2.5-VL-7B-Instruct, and Qwen2.5-VL-32B-Instruct \cite{liu2023visual, liu2024improved, bai2025qwen2}. These experiments follow the same experiment settings described in Section Experiment. The results for LLaVA-series are reported in Table \ref{tab:supp_llava} and Table \ref{tab:supp_llava-next}, while the results for Qwen-series are in Table \ref{tab:supp_qwen} and \ref{tab:qwen_32B}. When implemented on a model with a different size, HiPrune maintains its overall performance and shows a trend similar to a smaller one reported in our paper. Notably, for SQA and VizWiz, HiPrune acquires results even slightly better than baseline under some settings. HiPrune is model-agnostic and easy to deploy on other VLMs, which is among our future works.

\paragraph{Text-Dominant Results}
To evaluate the accuracy performance under text-dominant tasks, we test our method with DocVQA \cite{mathew2021docvqa} and present the results in Fig. \ref{fig:docvqa}. HiPrune and HiPrune++ surpass most comparisons under most token budgets, demonstrating robustness under text-dominant scenarios.

\subsection{Ablation Studies}
\label{sec:extended_ablation}
\paragraph{Hyperparameter Sentivity.}
HiPrune depends on two hyperparameters: the object layer $l$ and the object proportion $\alpha$. The former is set by the average pairwise distance of the top-attention token set, while the latter is set manually. As Fig. \ref{fig:hyperparam} shows, there does not exist an optimal point regarding $\alpha$, and the difference between various $\alpha$ is trivial, so we simply set it to 0.1 in all our settings. However, for different $l$, the optimal results emerge at layer 9, which is exactly the changing point of average pairwise distance in Fig. \ref{fig:layer_dist}. We hypothesize that this point features the most concentrated information and overlaps well with objects.

\paragraph{Buffer Selection Scheme.} 
Buffer tokens are tokens neighbouring anchor tokens. To better depict the positional relation between buffer tokens and anchor tokens, we provide a simple diagram in Fig. \ref{fig:buffer_selection}. Results for these schemes are reported in Table \ref{tab:buffer_selection}. When the number of buffers around one anchor surpasses 4, the performance stays stable. The buffers are introduced to mitigate misselection caused by noise in the attention map; therefore, theoretically, as long as their coverage size is sufficiently wide, the exact shape does not make a significant difference.

\section{Visualization of Attention Evolution}
As shown in Fig. \ref{fig:supp_tsne_all}, the high-attention tokens in the input layer and the output layer (last but one in LLaVA) obey distinct distributions. In these examples, high-attention tokens in the shallow layer distribute uniformly across the embedding space, while they cluster in the output layer. During the shift, the middle layers show a transitional status covering every cluster. These examples indicate that CLIP encodes images in a continuous and gradual way, forming a hierarchical representation inside the vision encoder.

\section{Visualizations on Retained Tokens}
\label{sec:supp_vis}
\begin{figure*}[t]
    \centering
    \includegraphics[width=\linewidth]{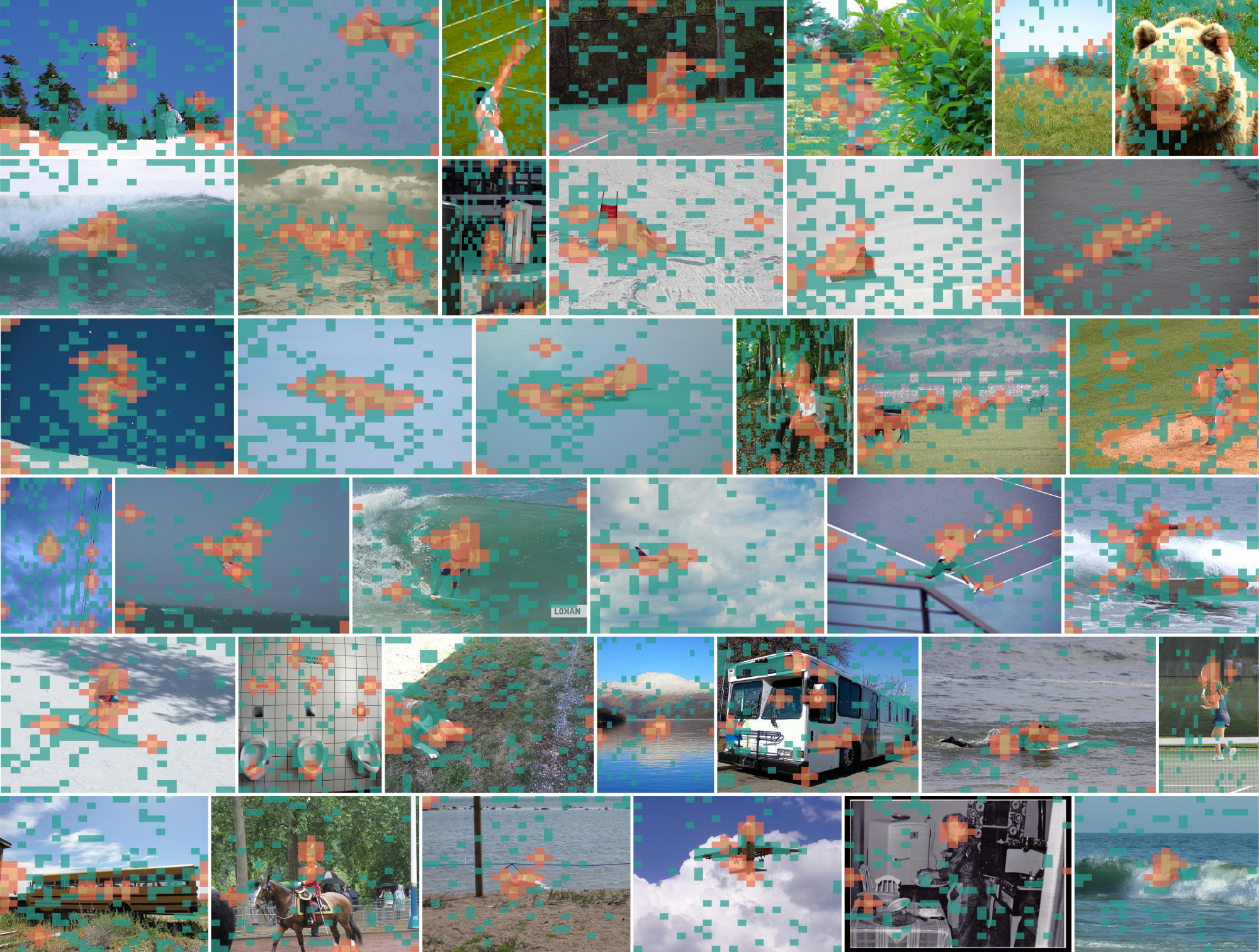}
    \caption{Visualization on tokens retained by HiPrune. The images are randomly chosen from the COCO val2017 set \cite{lin2014microsoft}. Anchor tokens are in yellow, buffer tokens are in red, and register tokens are in cyan.}
    \label{fig:supp_retained_tokens}
\end{figure*}

\begin{figure*}
    \centering
    \begin{minipage}{0.93\linewidth}
        \centering
        \includegraphics[width=\linewidth]{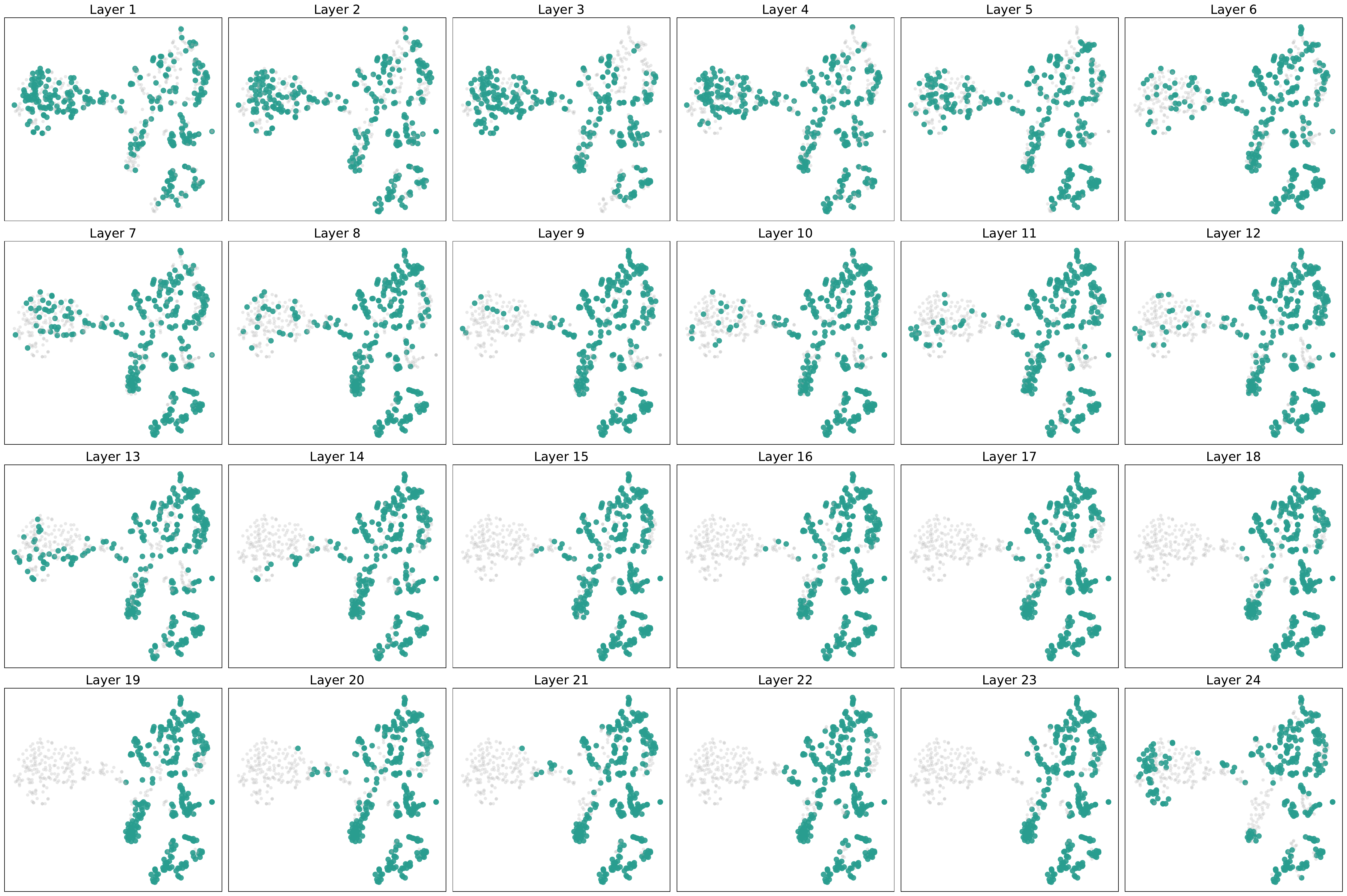}
    \end{minipage}
    \begin{minipage}{0.93\linewidth}
        \centering
        \includegraphics[width=\linewidth]{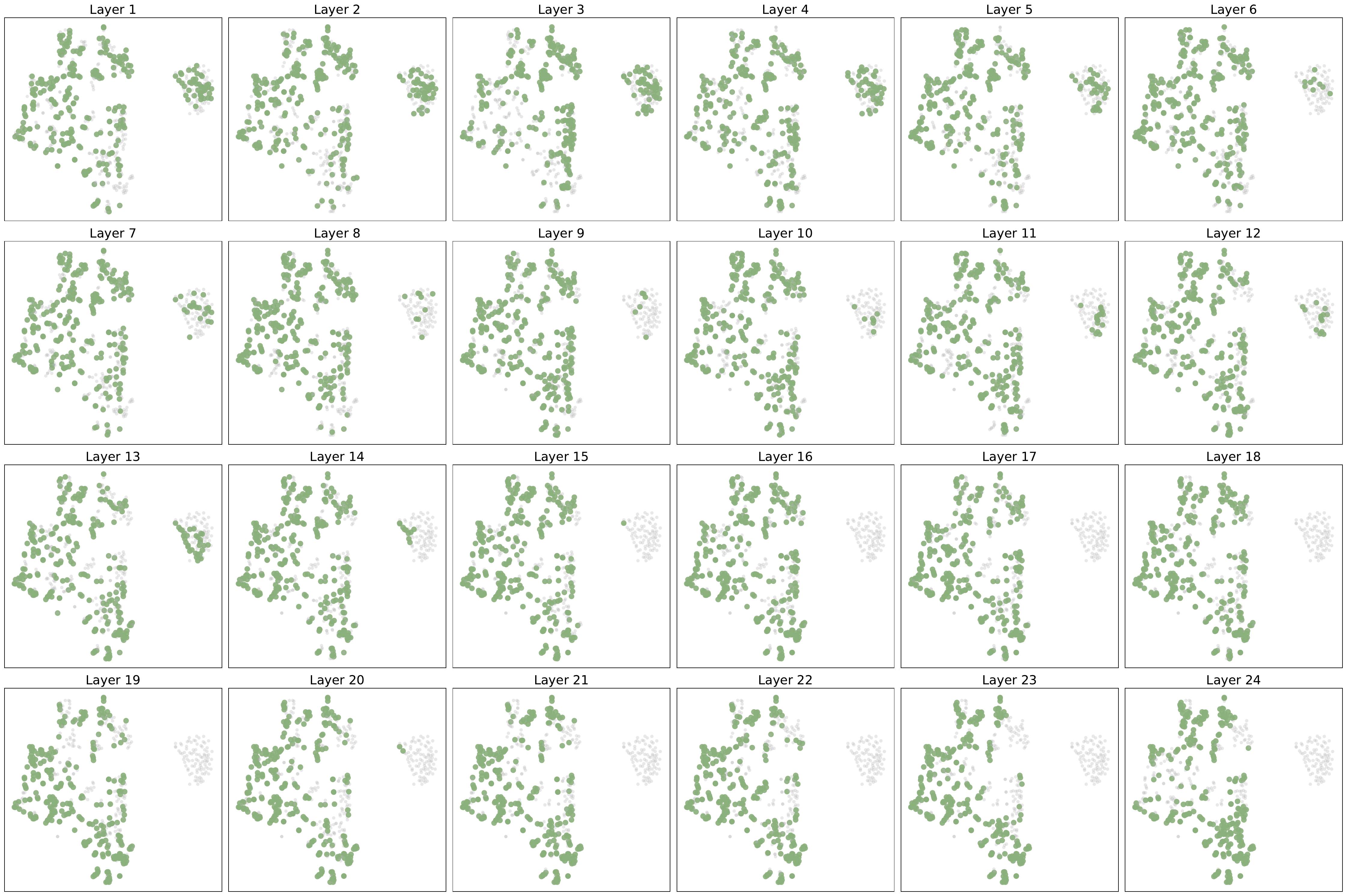}
    \end{minipage}
    \caption{Two examples of t-SNE visualization on visual tokens that receive top 50\% attention from different CLIP layers.}
    \label{fig:supp_tsne_all}
\end{figure*}

We provide extended examples on retained tokens in Fig. \ref{fig:supp_retained_tokens}. We can see that the anchor tokens and buffer tokens contain the main body of the image, e.g., the player, the person, and the animal, which are crucial for image understanding. The register tokens seem less image-aligned, but cover the whole image sufficiently, indicating they encode global information. The reason is discussed in the section Experiment, and this visualization further confirms our theory.

\section{Visualizations on All Layers' Attention}
\begin{figure*}[t]
    \centering
    \includegraphics[width=\linewidth]{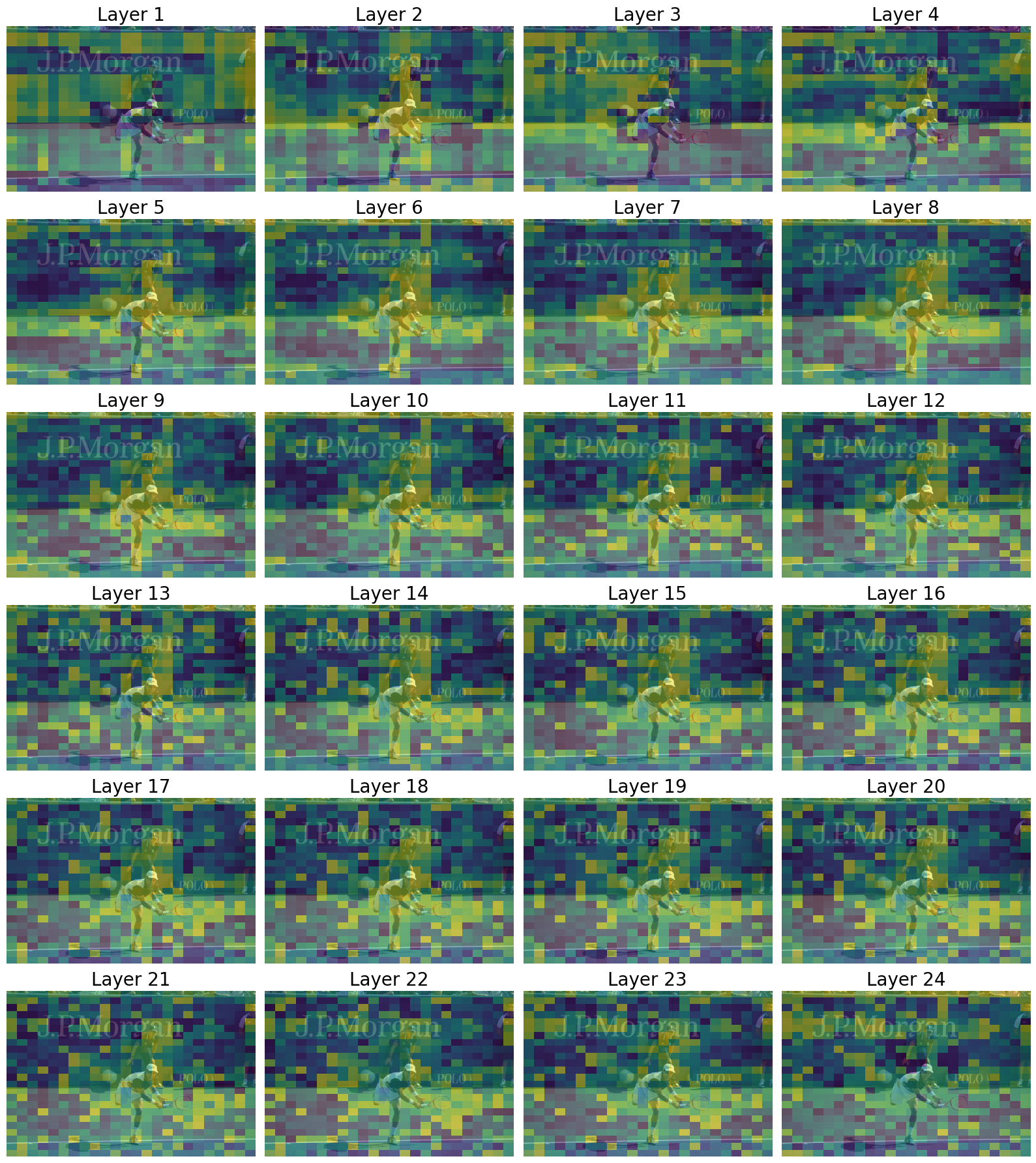}
    \caption{Visualization of all layers' attention in the CLIP model.}
    \label{fig:supp_layer_1}
\end{figure*}

We provide two examples of all layers' attention distribution in Fig. \ref{fig:supp_layer_1}. In the middle layers, the highlighted areas mainly overlap the person, which is the main object of the image, and agrees with the human perception order, as we also first notice the person in the image.

\end{document}